\documentclass{isprs} 
\usepackage{setspace}
\usepackage{geometry} 
\usepackage{epstopdf}
\usepackage[labelsep=period]{caption}  
\usepackage[british]{babel} 
\usepackage{amssymb}
\usepackage{amsmath}
\usepackage[font=scriptsize]{subfig}
\usepackage{color}
\usepackage{multirow}
\usepackage{multicol}
\usepackage{array}

\definecolor{pinegreen}{RGB}{1, 121, 111}

\newcommand{\extended}[1]{\textcolor{black}{{#1}}}
\newcommand{\newtext}[1]{\textcolor{black}{{#1}}}

\newcolumntype{P}[1]{>{\centering\arraybackslash}p{#1}}

\begin{document}
	
	\title{Unsupervised Automatic Building Extraction Using Active Contour Model on Unregistered Optical Imagery and Airborne LiDAR Data}
	
	%

	\author{
		T. H. Nguyen\textsuperscript{a,b,}\thanks{Corresponding author} , %
		S. Daniel\textsuperscript{b}, D. Gu\'{e}riot\textsuperscript{a}, C. Sint\`{e}s\textsuperscript{a}, J.-M. Le Caillec\textsuperscript{a}}
	
	\address{
		\textsuperscript{a }IMT Atlantique, Lab-STICC, UMR CNRS 6285, F-29238 Brest, France -  \\
		(thanh.nguyen, didier.gueriot, christophe.sintes, jm.lecaillec)@imt-atlantique.fr\\
		\textsuperscript{b }Universit\'{e} Laval, Qu\'{e}bec City, QC G1V 0A6, Canada - sylvie.daniel@scg.ulaval.ca
	}


	\icwg{II/III}   
	
	%
	
	\abstract
	{
		Automatic extraction of buildings in urban scenes has become a subject of growing interest in the domain of photogrammetry and remote sensing, 
		particularly with the emergence of LiDAR systems since mid-1990s. However, in reality, this task is still very challenging due to the complexity of building size and shape, as well as its surrounding environment.
		Active contour model, colloquially called snake model, which has been extensively used in many applications in computer vision and image processing, has also been applied to extract buildings from aerial/satellite imagery. Motivated by the limitations of existing snake models dedicated to the  building extraction, this paper presents an unsupervised and  automatic snake model to extract buildings using optical imagery and an unregistered airborne LiDAR dataset, without manual initial points or training data. The proposed method is shown to be capable of extracting buildings with varying color from complex environments, and yielding high overall accuracy.
	}

	
	\maketitle
	
	
	\section{CONTEXT}\label{sec:intro}
	Automatic extraction of buildings from aerial/satellite imagery in urban and residential scenes has become a subject of growing interest in the domain of photogrammetry and remote sensing. Indeed, a large number of building detection and extraction techniques have been reported over the last few decades, particularly with the emergence of LiDAR (Light Detection and Ranging) systems. 
	For instance, \cite{Rottensteiner2003174} proposed a building detection method exploiting primarily LiDAR data while removing vegetation using imagery data. \cite{sohn2007data} focused on exploiting the synergy of IKONOS multispectral imagery combined with a hierarchical segmentation of a LiDAR digital elevation model (DEM) to extract buildings. Another method of building detection was proposed by \cite{awrangjeb2010automatic} based on building masks obtained from LiDAR and multispectral imagery. These methods using both complementary sources, namely LiDAR data and optical imagery, achieve better building extraction results. Other building detection and extraction methods utilize only LiDAR data such as \cite{khoshelham2013segment,zhang2013svm}. They involve segmentation algorithms and classification using attributes such as building size, shape, height and PCA (Principal Component Analysis) features. However, these approaches usually face problem of misclassification of vegetation as buildings \cite{zhang2017advances}.
	
	Originally introduced by \cite{kass1988snakes}, snake model has been studied for building extraction from urban area using aerial and satellite imagery. While \cite{guo2002snake} used snake model with balloon force to extract buildings, \cite{peng2005improved} focused in increasing stability of snake convergence.  \cite{kabolizade2010improved} proposed to improve the model with imagery data coupled with a Digital Surface Model (DSM) generated from LiDAR data.  
	This improvement involves 
	using two energy terms based on the  variances of height and gray-level between snake points. Hence, the height variance energy term requires height information for every pixel of the image, in other words, the DSM must be of same size and resolution as the imagery data. This can be problematic since LiDAR dataset does not often have a spatial resolution as high as aerial imagery, and height data interpolation may be unreliable. 
	\cite{fazan2013rectilinear} proposed another approach based on exhaustive searches of rectilinear building corners from the image, optimized by dynamic programing. However, this method depends heavily on initial points to have decent results. 
	\cite{ahmadi2010automatic} proposed an area-based geometrical snake model to detect building boundaries without height information or initial points. Nevertheless, this method requires \textit{a priori} sampled gray-levels of buildings and grounds (i.e. training data) to attract the snakes toward desired buildings. It also may not work well with building roofs with varying gray levels. 
	
	To summarize, the main common issues of snake models in building boundary extraction are:
	\vspace{-0.2cm}
	\begin{itemize}
		\itemsep-0.4em
		\item Sensitivity to noise and image details;
		\item Dependence on initial points or training data;
		\item Weak convergence to building corners; 
		\item Snake's convergence sensitivity to its number of points and weighting parameters.
	\end{itemize}
	
	Motivated by the limitations of existing snake models listed above, this paper presents an unsupervised  snake model to extract buildings using optical imagery and an unregistered airborne LiDAR dataset, without manual initial points or training data. \newtext{Instead, we propose to initialize our proposed snake model with projected LiDAR building boundary points. The movement of snake model is then governed by an additional energy term related to its similarity to the LiDAR building boundary.}  
	Our method is also able to extract buildings with varying color, from complex environments.
	
	\extended{The remainder of this paper is organized as follows. Section \ref{sec:method} is devoted to the description of the proposed method. 
		Then, experimental results are presented in Section \ref{sec:result}. Finally, Section \ref{sec:conclusion} provides conclusions and perspectives of this work.}
	
	\section{METHOD}\label{sec:method}
	
\begin{figure}[t]
		\centering
		\includegraphics[trim=0 18cm 8cm 0,clip,width=\linewidth]{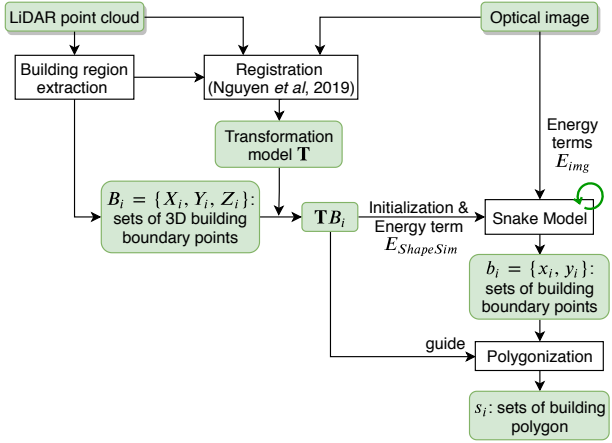}
		\caption{Flowchart of the proposed method.}
		\label{fig:workflow}
	\end{figure}

	\extended{The novelty of our approach resides in an unsupervised method that carries out effectively building extraction from urban scene. This method is based on a snake model initialized and enhanced by integrating with LiDAR data, followed by an improved building polygonization. 
		Presented by Fig. \ref{fig:workflow}, the proposed method is composed of three  parts: \textit{(i)} determination of initial points involving the registration of datasets, \textit{(ii)} the snake model, and \textit{(iii)} the improved building boundary polygonization.}

	\subsection{Determination of initial points}
	\subsubsection{Registration}
	This research study involves an airborne LiDAR data set which has been acquired independently of the optical imagery, i.e. two different surveys, done at different time, with different platforms. Such a context yields spatial discrepancies between data sets that provide the snake model with wrong initial points. Therefore, a registration has been carried out beforehand \cite{nguyen:hal-02075341}. Indeed, such a registration is substantially important to a building extraction procedure, since it involves many problems exemplified by the work of \cite{gilani2016automatic}. 
	
	The registration consists in extracting building regions, on one hand, from the LiDAR point cloud, and on the other hand, from the optical image using a mean-shift segmentation. \extended{Extracted building segments are then matched using the Graph Transformation Matching algorithm proposed by \cite{aguilar2009robust}. The resulting pairwise segments establish a set of correspondences between the two data sets.} They are used to estimate a transformation model using the Gold Standard algorithm \cite[p. 187]{hartley2003multiple}. This transformation model is used to register the image and LiDAR data sets with an accuracy good enough to get relevant snake initial points.
	
	
	
	\subsubsection{Building region extraction from LiDAR point cloud}
	
\begin{figure}[t]
		\centering
		\includegraphics[trim=0cm 19.5cm 10.5cm 0,clip,width=0.95\linewidth]{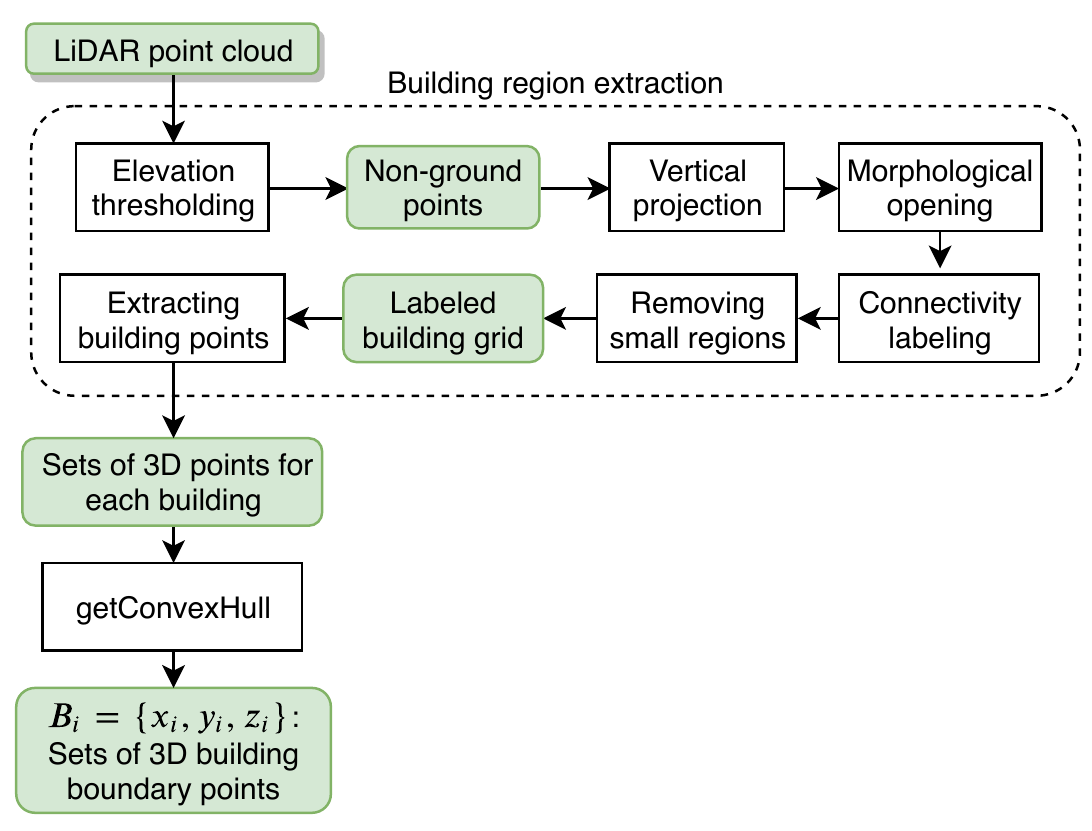}
		\caption{Building region extraction from LiDAR point cloud.}
		\label{fig:lidar_BE}
	\end{figure}

	Presented by Fig. \ref{fig:lidar_BE}, the extraction of building regions from LiDAR point cloud is carried out through a series of steps. First, non-ground points are separated from ground points using an elevation thresholding. The threshold is set as follows, $ T_e=\mathrm{mean}(z_G) + \max\{2.5, \mathrm{std}(z_G)\} $, where $z_G$ denotes the elevation of ground points. All non-ground points are then vertically projected onto the plan $ z=0 $. A raster grid representing these projected points is created. The resolution of the grid is set according to the LiDAR point cloud density in order to avoid null-valued pixels. 
	A binary grid of same resolution is also generated. Its cell value is set to 1 or 0 according to the presence or absence of projected non-ground points in the cell  (\textquoteleft1\textquoteright: presence, \textquoteleft0\textquoteright: absence).
	A morphological opening operator is then applied to remove small artifacts on the binary grid. Remaining grid cells with value set to 1 are grouped into labeled segments based on their connectivity. Next, small segments (e.g. smaller than 10 m$^2 $) are removed. The resulting grid consists of a number of relatively large labeled segments that relate to buildings. These segments are then used to select the building points in the LiDAR point cloud.
	A convex hull  is calculated on each set of 3D building points, yielding a set of boundary points, denoted by $ B_i $. These points are then projected onto the optical image using the transformation model $ \mathbf{T} $. They will be used as the initial points of the snake model for each building.
	
	\subsection{Snake models for building boundary extraction}

	
	\subsubsection{Traditional snake models}
	An active contour, or a snake is a dynamic curve $ \mathbf{x}(s)=(x(s),y(s)) $, where $ s \in [0,1]$ is the normalized arc length, defined within an image domain that is deformable under the influence of internal and external forces \cite{xu1997gradient}. Mathematically, the behaviors of the snake are governed by an energy function, which is defined as follows,
	\begin{equation}\label{eq:snake1}
	\begin{array}{l}
	E_{\mathrm{snake}}=\int_{0}^{1}(E_{\mathrm{int}}(\mathbf{x}(s)) + E_{\mathrm{ext}}(\mathbf{x}(s))) ds\\[7pt]
	E_{\mathrm{int}}(\mathbf{x}(s))=\dfrac{1}{2} \left(\alpha |\mathbf{x}'(s)|^2 + \beta|\mathbf{x}''(s)|^2\right) \\[7pt]
	E_{\mathrm{ext}}(\mathbf{x}(s))=E_{\mathrm{img}}(\mathbf{x}(s))+E_{\mathrm{con}}(\mathbf{x}(s))
	\end{array}
	\end{equation}
	where  $ E_{\mathrm{int}} $ and $ E_{\mathrm{ext}} $, respectively, represent the internal and external energy terms. 
	The internal energy relates to the tension (the amount of stretch) and the rigidity (the amount of curvature) of the snake, respectively controlled by weighting parameters 
	$ \alpha $ and $ \beta $. 
	$ \mathbf{x}'(s) $ and $ \mathbf{x}''(s) $ denote the first and second derivatives of $ \mathbf{x}(s) $ with respect to $ s $. 
	The external energy $ E_{\mathrm{ext}} $ is composed of the forces due to the image itself $ E_{\mathrm{img}} $, and other constraint forces introduced by the users $ E_{\mathrm{con}} $, e.g. inflation force introduced by balloon model \cite{cohen1991active}.
	
	The external energy of an image $ I $ related to its salient features i.e. lines, edges and terminations (i.e. line end-points, corners) can be generally formulated as follows,
	\begin{equation}
	E_{\mathrm{img}}=w_{line}E_{line}+w_{edge}E_{edge}+w_{term}E_{term}
	\end{equation}
	where $ w_{line}, w_{edge}, w_{term} $ are the weights of the respective salient features. Mathematical formulation of these energy terms 	is described in \cite{kass1988snakes}. 


	\extended{A snake that minimizes $ E_{\mathrm{snake}}$ must satisfy the Euler equation
		\begin{equation}\label{eq:euler}
		\alpha\mathbf{x}''(s) - \beta\mathbf{x}''''(s) - \nabla E_{\mathrm{ext}}=0
		\end{equation}
		In order to solve \eqref{eq:euler}, $  \mathbf{x} $ is regarded as a function of time $ t $ as well as of $ s $. Then, the partial derivative of $   \mathbf{x} $ with respect to $ t $ is then set equal to the left hand side of \eqref{eq:euler}, as follows,
		\begin{equation}\label{eq:snake_iter}
		\mathbf{x}_t(s,t)=\alpha\mathbf{x}''(s) - \beta\mathbf{x}''''(s) - \nabla E_{\mathrm{ext}}
		\end{equation}
		When $ \mathbf{x}(s,t) $ stabilizes, the partial derivative term $ \mathbf{x}_t(s,t) $ vanishes and we obtain a solution for \eqref{eq:euler}. This dynamic equation can also be viewed as a gradient descent algorithm designed to solve \eqref{eq:snake1}. A numerical solution to \eqref{eq:snake_iter} can be found by discretizing the equation and solving the discrete system iteratively \cite{kass1988snakes}.}

	\subsubsection{Gradient vector flow} (GVF) is proposed by \cite{xu1997gradient} to allow more flexible  initialization of snake and encourage convergence to boundary concavities, as well as improving the model's robustness versus image noise. 
	GVF field is defined as the vector field $ \mathbf{v}(x,y) = (u(x,y), v(x,y)) $ that minimizes the energy functional
	\begin{equation}\label{gvf1}
	E_{\mathrm{GVF}} = \int\int\mu (u_x^2+u_y^2+v_x^2+v_y^2) + |\nabla f|^2|\mathbf{v}-\nabla f|^2dxdy
	\end{equation}
	with $ \mu $ is a controllable smoothing term, and $ f $ represents external forces from Eq. \ref{eq:euler}, i.e. $ f(x,y)=-E_\mathrm{ext} $.
	Using the \textit{calculus of variations} \cite{courant2008methods}, the GVF can be found by solving:
	\begin{equation}\label{gvf2}
	\begin{array}{l}
	\mu\nabla^2u-(u-f_x)(f_x^2+f_y^2)=0\\[5pt]
	\mu\nabla^2v-(v-f_y)(f_x^2+f_y^2)=0
	\end{array}
	\end{equation}
	where $ \nabla^2 $ is the Laplacian operator. The Euler equations \eqref{gvf2} can be solved by regarding $ u $ and $ v $ as functions of time and solving
	\begin{equation}
	\begin{array}{l}
	u_t(x,y,t)=\mu\nabla^2u(x,y,t)- \\ [3pt]
	\textnormal{~~~~~~~~~~} [u(x,y,t)-f_x(x,y)]\cdot[f_x(x,y)^2+f_y(x,y)^2]\\[5pt]
	v_t(x,y,t)=\mu\nabla^2v(x,y,t)-\\[3pt]
	\textnormal{~~~~~~~~~~} [v(x,y,t)-f_y(x,y)]\cdot[f_x(x,y)^2+f_y(x,y)^2]\\
	\end{array}
	\end{equation}
	Once computed, $ \mathbf{v}(x,y) $ will replace the potential force $ -\nabla E_{\mathrm{ext}} $ in the dynamic equation \eqref{eq:euler}, yielding
	\begin{equation}
	\mathbf{x}_t(s,t)=\alpha\mathbf{x}''(s) - \beta\mathbf{x}''''(s) + \mathbf{v}
	\end{equation}
	This equation is solved similarly as the traditional snake model, i.e. by discretization and iterative solution. The parametric curve solving the above dynamic equation is thus called a GVF snake. 
	
	\subsubsection{Proposed snake model}
	In this paper, we propose adding a new energy term as a constrained force (i.e. $ E_{\mathrm{con}} $), which is calculated based on the similarity between the shape formed by snake points and the projected LiDAR building boundary, as follows,
	\begin{equation}\label{DF}
	E_{\mathrm{ShapeSim}}(\mathbf{x})=1-\exp\left(-\dfrac{d^2_H (\mathbf{x},\mathbf{T}{B_i})}{\delta}\right)
	\end{equation}
	where $ d_H $ denotes the Hausdorff distance, $ \delta $ is the scale factor set accordingly to the size of the building, and $ \mathbf{x} $ is the current snake. $ B_i $ represents the 3D building boundary extracted from LiDAR data, and  $ \mathbf{T}B_i $ is its projection onto the image using the transformation model $ \mathbf{T}$ provided by the registration. 
	The  role of this energy term is to encourage the snake to maintain the same shape as the building boundary extracted from the LiDAR data while being attracted by the image salient features and under the influence of GVF fields.
	
\begin{figure}[t]
		\centering
		\includegraphics[trim=4.5cm 1.5cm 3cm 0.5cm,clip,width=\linewidth]{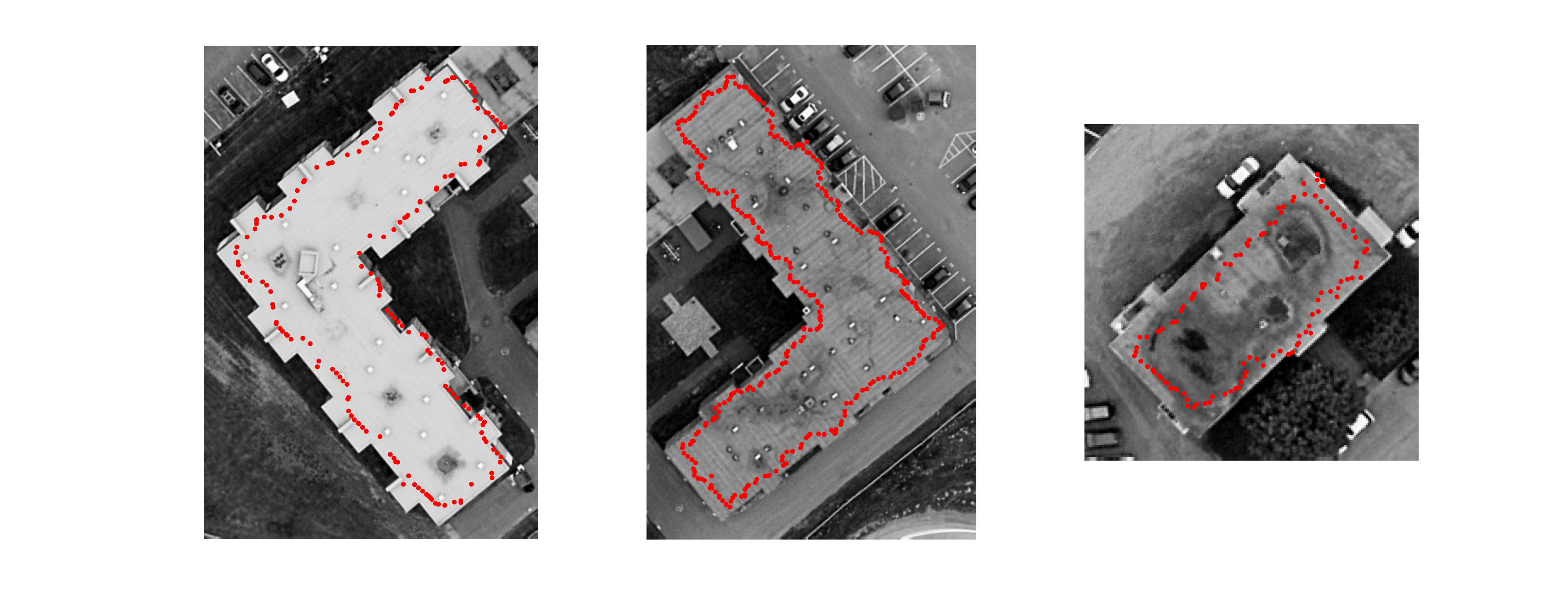}
		\caption{3D building boundary points projected onto the optical image, i.e. $ \mathbf{T}B_i $, which are used to initialize the snake model and for the energy term $ E_{\mathrm{ShapeSim}} $.}
		\label{fig:initial points}
	\end{figure}

	In our approach, the projected LiDAR building boundaries, denoted by $ \mathbf{T}B_i $, have an important role. They provide the snake algorithm with relevant initial points, and they are used in the energy term $ E_{\mathrm{ShapeSim}} $ and the polygonization. Fig. \ref{fig:initial points} shows three examples of 3D building boundary points projected onto the optical image. 
	
	Up to this point, the proposed snake model has involved a number of parameters, such as $ \alpha, \beta $ (from $ E_\mathrm{int} $), $ w_{line}, w_{edge}, w_{term} $ and $ \sigma $ (from $ E_\mathrm{img} $), $ \mu $ (from GVF) and lastly $ \delta $ (from $ E_{\mathrm{ShapeSim}} $). To the best of our knowledge, the know-how to set these parameters for the snake to extract buildings from an imagery dataset still remains unresolved. This has driven many existing works to perform a trial-and-error approach to determine them \cite{peng2005improved,kabolizade2010improved,ahmadi2010automatic}. 
	Similar approach is applied for our snake model, empirically setting $ \alpha=\beta=0.01,  w_{line}=0.04, w_{edge}=2, w_{term}=0.01 $, with standard deviation $ \sigma=10 $ used for  image smoothing in $ E_\mathrm{img} $, $ \mu=0.2 $ and lastly $ \delta=50 $.
	
	
	%
	\subsection{Polygonization of building boundary}
	The polygonization step is then applied to each set of building boundary points extracted with the snake model, i.e. $ b_i $.
	\newtext{It involves formulating these boundary points into regular building polygons. Such polygon usually consists of several parallel and perpendicular straight line segments.}
	Many researches have addressed this topic for various purposes, such as  building boundary \cite{dutter2007generalization},  cartography \cite{samsonov2017shape}, and classification maps \cite{maggiori2017polygonization}.
	
	In this paper, we modified the method proposed by \cite{dutter2007generalization} which polygonizes a building into three levels of shape: rectangular; Z-, T- or L-shape; U-shape. The original method starts with a Minimum Bounding Rectangle (MBR) \cite{freeman1975determining} containing all snake points, and the edges of building polygon are set to be parallel to the MBR edges. This setting may bias the polygonization result and eventually lead to errors in building orientation. In our proposed approach, instead of detecting the MBR containing snake points, we detect and use the MBR of the projected LiDAR building boundary points which yields more reliable building orientation.
	
\begin{figure}[t]
		\centering
		\subfloat[Comparison of polygonization algorithms]{\includegraphics[trim=3.5cm 0.5cm 1.4cm 0.5cm,clip,width=0.5\linewidth]{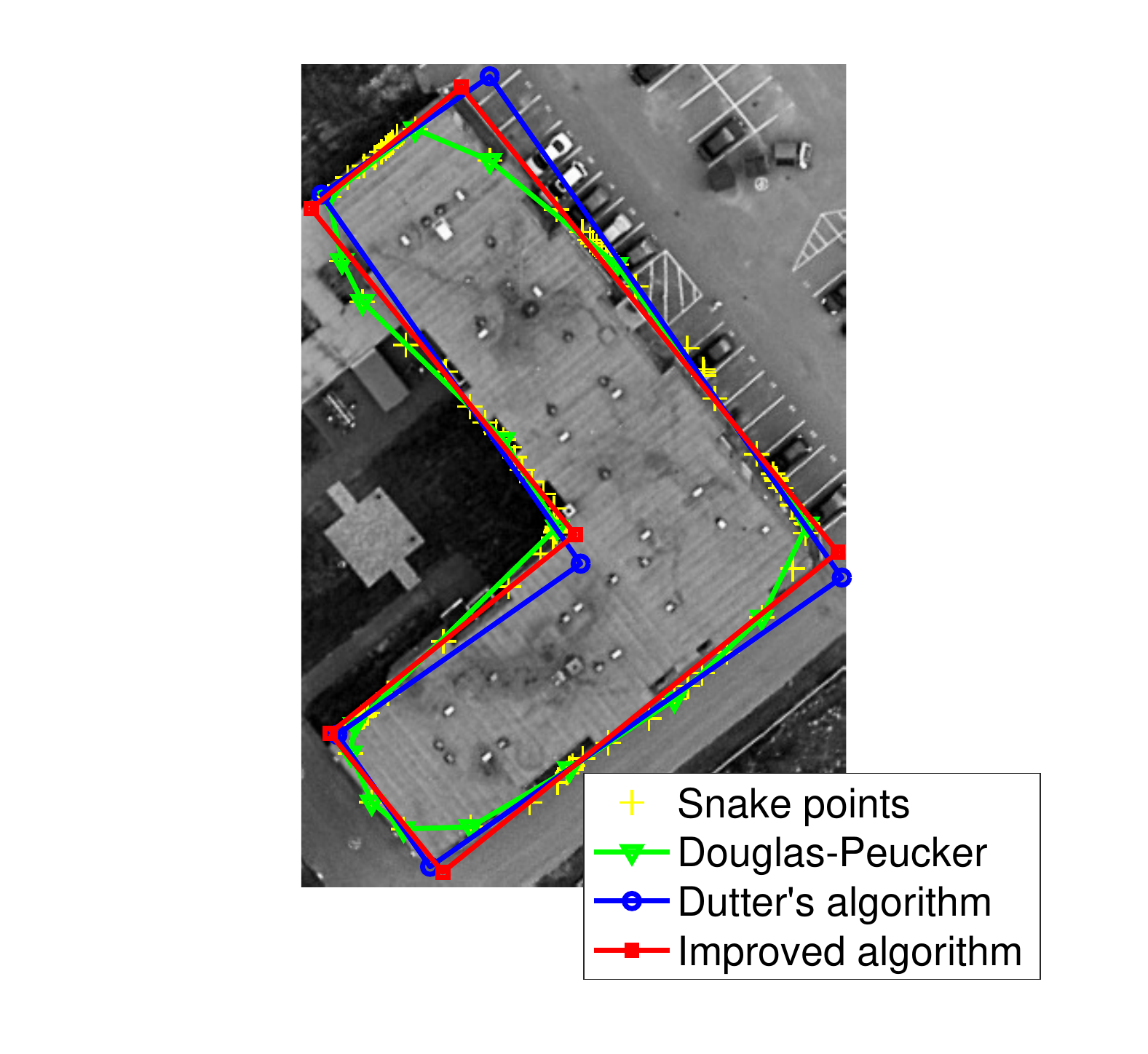}\label{fig:compare_polygon}}\hspace{0.05cm}
		\subfloat[Building orientation error]{
			\scalebox{0.8}{%
				\begin{tabular}[b]{cc}
					\hline
					Algorithm & Error \\
					\hline
					\multirow{2}{2cm}{\cite{douglas1973algorithms}} & \multirow{2}{*}{-} \\
					& \\
					\hline
					\cite{dutter2007generalization} &$ 4.2^\circ $\\
					\hline
					Improved algorithm & $ 0.73^\circ  $\\
					\hline
				\end{tabular}
				\label{tab:error}
			}
		}
		\caption{Polygonization of the proposed snake model on an L-shaped building.}
		\label{fig:polygonization}
	\end{figure}

	A comparison of building boundary polygonization methods has been carried out. It involves our approach, Dutter's original polygonization \cite{dutter2007generalization} and Douglas-Peucker line simplification algorithm \cite{douglas1973algorithms}. Results are provided by Fig. \ref{fig:polygonization}, which underline the significant reduction of building orientation error yielded by our approach. Since Douglas-Peucker algorithm only reduces the number of points of a boundary, it does not yield a building orientation.
	
	\section{Results}\label{sec:result}
	This section is devoted to the assessments of the snake model  and  the overall building extraction performance. 
	
	\subsection{Evaluation metrics}\label{ssec:metric}
	In order to assess performance of a building extraction method with respect to a ground truth, the following metrics are used: 
	\vspace{-.25cm}
	\begin{enumerate}
		\itemsep0em
		\item Intersection over Union (IoU) \cite{jaccard1901etude}:
		\begin{equation}\label{eq:iou}
		\mathrm{IoU}=\dfrac{\mathcal{A}(E\cap R)}{\mathcal{A}(E\cup R)}\times 100\%
		\end{equation}
		where $ \mathcal{A}(\cdot) $ represents the area measurement. $ \mathrm{IoU} $ measures the ratio between the area of the intersection of an extracted building boundary $ E $ and the ground-truth building boundary $ R $, over the area of their union.
		
		Two related criteria, namely \textit{Completeness} ($ \mathrm{C}_p $) and \textit{Correctness} ($ \mathrm{C}_r $), are also used to facilitate the comparison of our approach with other works, e.g. \cite{fazan2013rectilinear,gilani2016automatic}. Full descriptions of these metrics can be found in the paper of  \cite{rutzinger2009comparison}.
		\begin{itemize}
			\itemsep0em
			\item Completeness, measured by the \textit{Recall} criterion: 
			\begin{equation}\label{cri:comp}
			\mathrm{C}_p = \dfrac{TP}{TP+FN}
			= \dfrac{\mathcal{A}(E\cap R)}{\mathcal{A}(R)}\times 100\%
			\end{equation}
			\item Correctness, measured by the \textit{Precision} criterion: 
			\begin{equation}\label{cri:corr}
			\mathrm{C}_r = \dfrac{TP}{TP+FP} = \dfrac{\mathcal{A}(E\cap R)}{\mathcal{A}(E)}\times 100\%
			\end{equation}
		\end{itemize}
		where $ TP $, $ FP $ and $ FN $ respectively denote the number of true positive (i.e. building pixels correctly identified as building pixels), false positive (i.e. non-building pixels identified as building pixels) and false negative pixels (i.e. building pixels not identified). 
		
		All three metrics IoU, $ \mathrm{C}_p $ and $ \mathrm{C}_r $ reach their best value at 100\% and worst at 0\%.
		
		While the IoU metric reflects the overall accuracy of a building extraction method according to a ground truth, $ \mathrm{C}_p $ represents the fraction of relevant identified building pixels over total number of actual building pixels,
		and $ \mathrm{C}_r $ represents the fraction of relevant identified building pixels among all identified pixels.
		
		\item Euclidean distance between centroids (EDC):
		\begin{equation}\label{eq:edc}
		\mathrm{EDC}=d(C_{E},C_{R})
		\end{equation}
		where $ d(\cdot) $ is the Euclidean distance, whereas $ C_E $ and $ C_R $, respectively, stand for the centroid coordinates of the extracted building and the ground truth building. 
		\item Dominant angle rotation error (DARE):
		\begin{equation}\label{eq:dare}
		\mathrm{DARE}=|\theta_{E}-\theta_{R}|
		\end{equation}
		where $ \theta_{E} $ and $ \theta_{R} $ represent the dominant angle of the extracted building and the ground truth building. The dominant angle of a polygonized building is determined according to its longest line segment. 
		
	\end{enumerate}
	
	\subsection{Data sets}
	\begin{figure*}[t]
		\centering
		\subfloat[]{\includegraphics[trim=2.5cm 1.5cm 2.5cm 0cm,clip,width=0.135\linewidth]{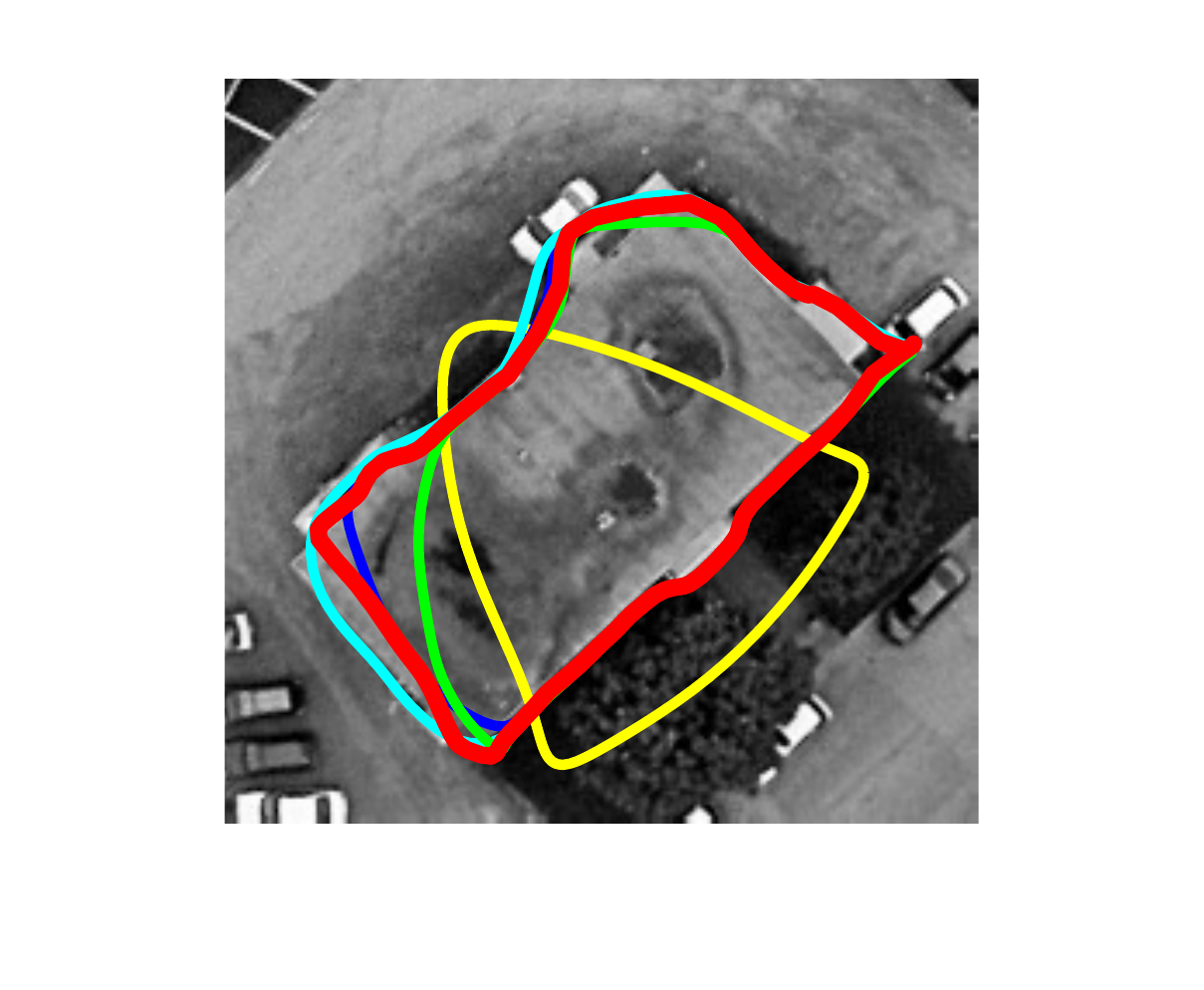}}\hspace{0.1cm}
		\subfloat[]{\includegraphics[trim=2.5cm 1.5cm 2.5cm 0.75cm,clip,height=4.25cm]{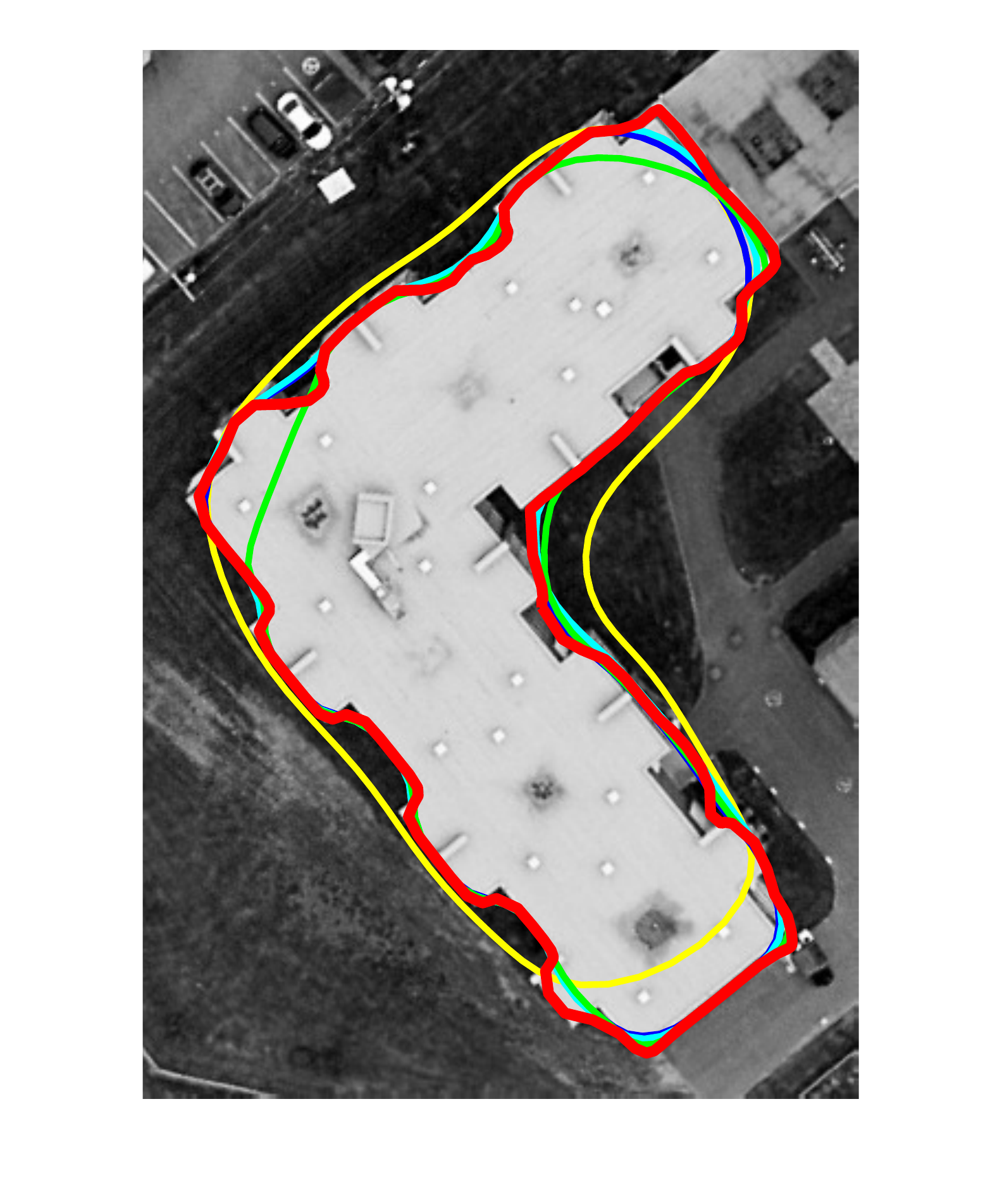}}\hspace{0.1cm}
		\subfloat[]{\includegraphics[trim=2cm 1.5cm 2.5cm 0.75cm,clip,height=4.25cm]{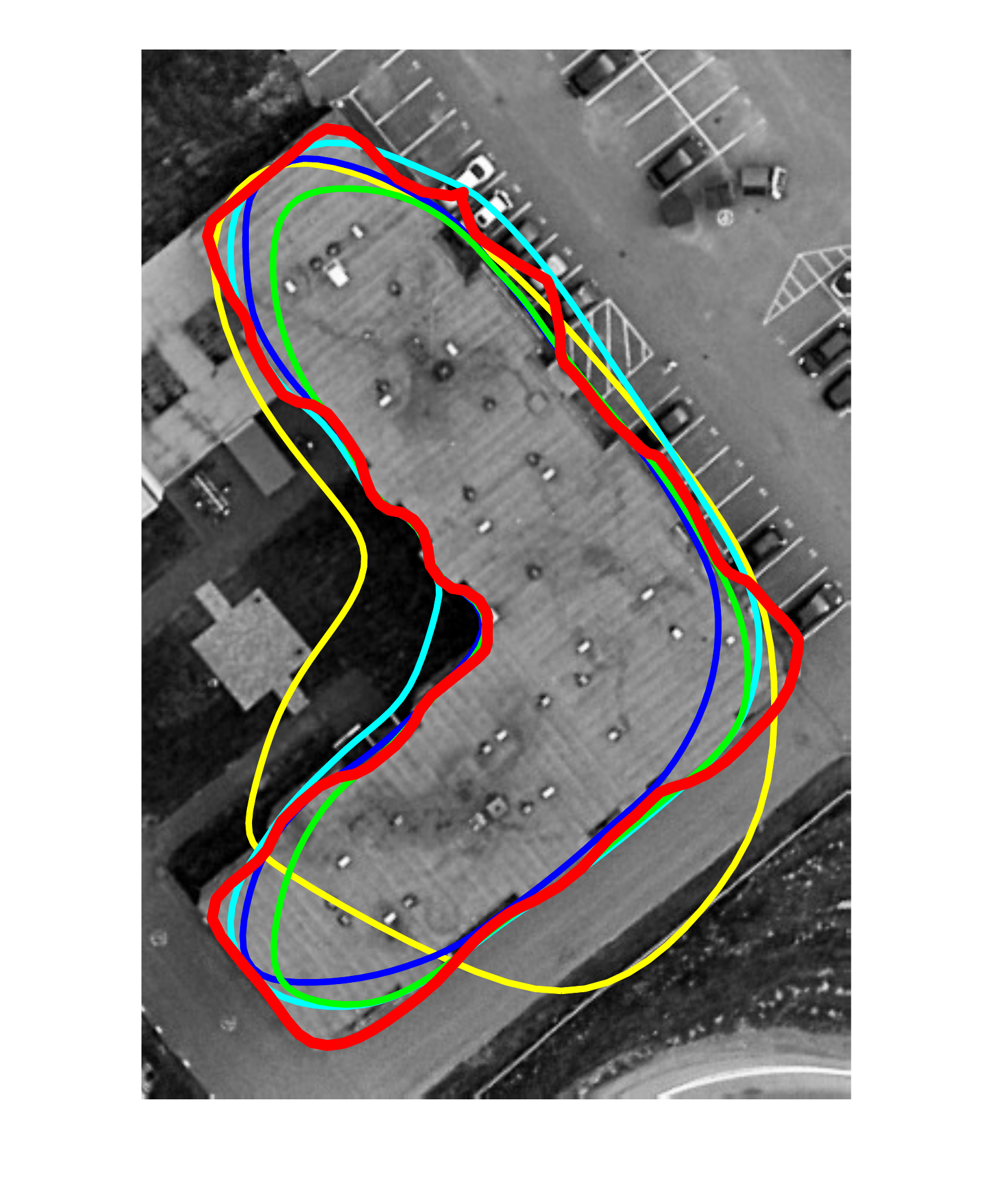}\label{sfig:c}}\hspace{0.1cm}
		\subfloat[]{\includegraphics[trim=1.75cm 1.75cm 1.75cm 1.5cm,clip,height=4.25cm]{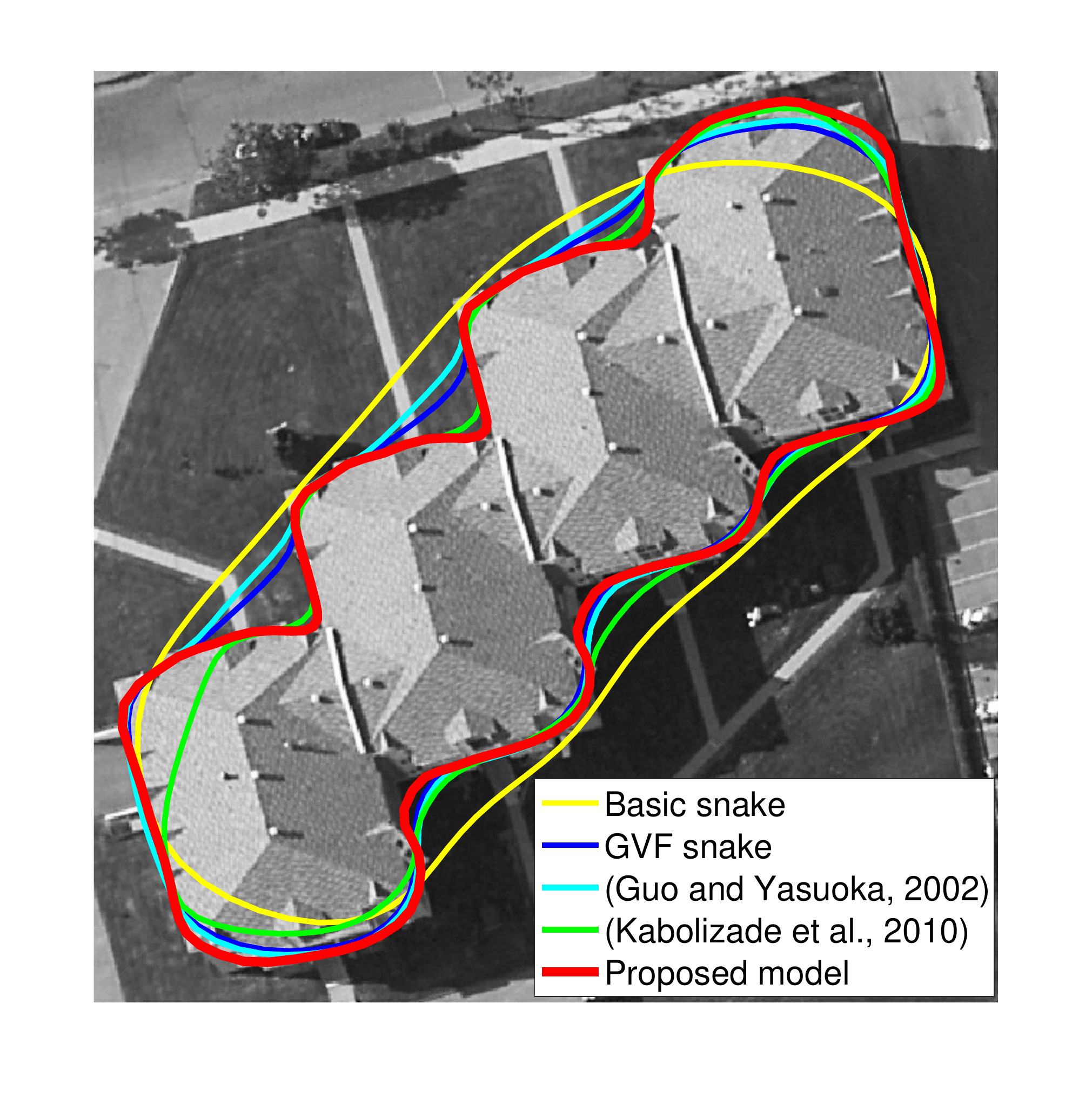}\label{sfig:d}}\hspace{0.2cm}
		\subfloat[]{\includegraphics[trim=3cm 1.2cm 3cm 1.5cm,clip,height=4.25cm]{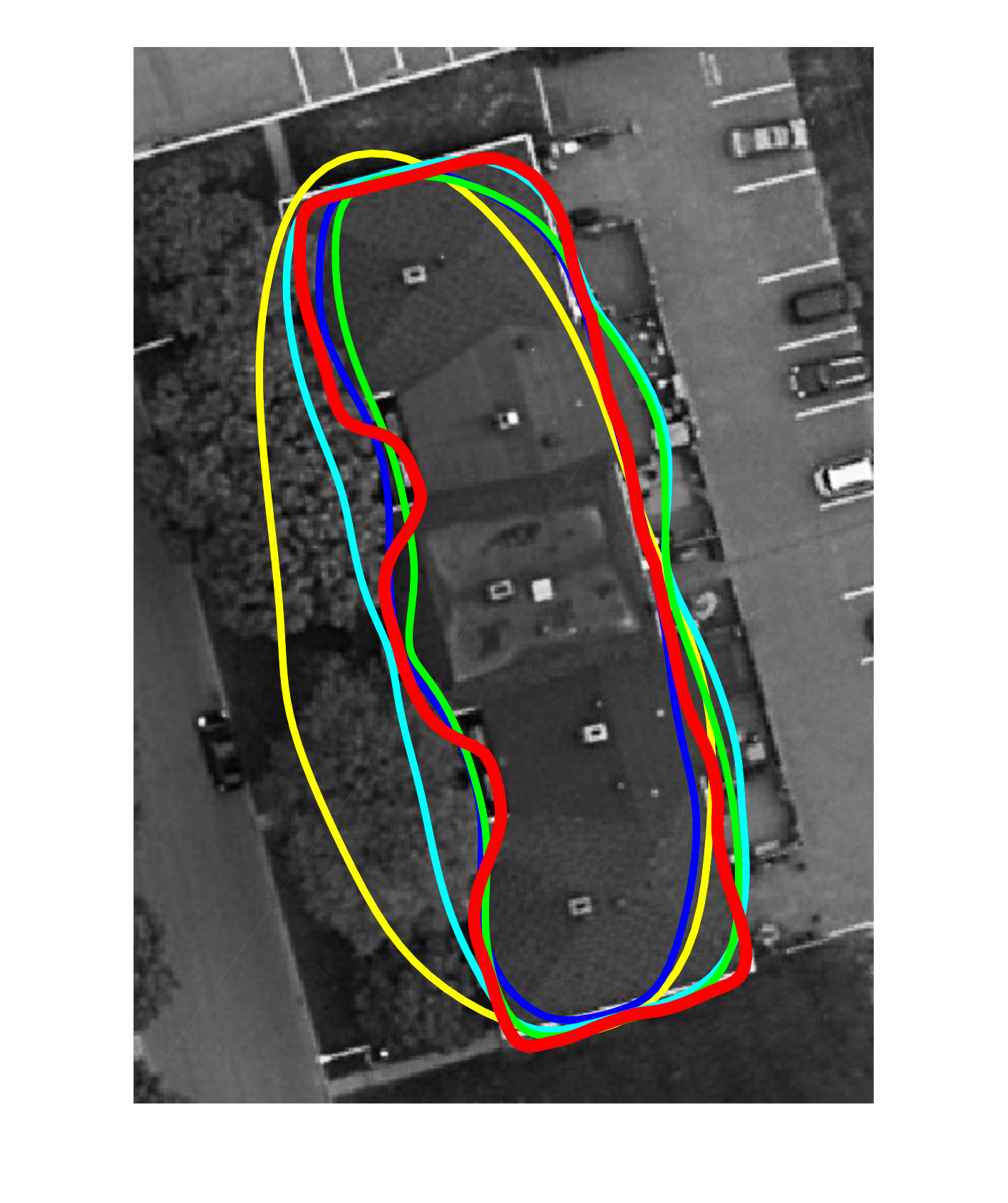}\label{sfig:e}}
		\vspace{-0.2cm}
		\caption{Results of snake models on simple and complex buildings. From left to right, (a) a rectangular building with varying color, then (b) an L-shaped building, (c) an L-shaped building with similar gray level to the background, (d) a gable roof building with its shadow, and (e) a gable roof building with similar gray level to the background.}
		\label{fig:compare_snakes2}
	\end{figure*}
	
	\subsubsection{Quebec City}
	The first data sets used for these assessments were acquired in Quebec City (QC, Canada) and are described in Table \ref{tab:datasets}. Boundary of twenty buildings on this test area are manually determined and used as ground truth. 
	\begin{table}[h]
		\centering
		\scalebox{0.85}{%
			\begin{tabular}{p{2.7cm}|P{2.85cm}|P{2.5cm}}
				\hline 			
				Data type & Aerial optical imagery & Airborne LiDAR \\\hline
				Spectral resolution & R, G, B, IR & 1064 nm \\\hline
				Spatial resolution/ & \multirow{2}{*}{15 cm} & \multirow{2}{*}{2 points/m$^2$}\\
				Point density & & \\\hline
				Acquisition time & June 2016  & Oct-Nov 2011  \\
				(season) & (summer) & (winter) \\\hline
				\multirow{2}{*}{Geometry/Properties} & Orthorectified & Multi-return (4) \\
				& Georeferenced & Classified \\\hline
				\multirow{2}{2.7cm}{Relative misalignment} & \multicolumn{2}{c}{1.41 m (before registration)}\\
				& \multicolumn{2}{c}{0.49 m (after registration)}\\
				\hline
		\end{tabular}}
		\caption{Description of Quebec City data set (LiDAR data \textcopyright Ville de Qu\'{e}bec, aerial imagery data \textcopyright Communaut\'{e} M\'{e}tropolitaine de Qu\'{e}bec).}
		\label{tab:datasets}
	\end{table}
	
	\subsubsection{Vaihingen}
	The proposed method is also tested using the ISPRS benchmark data set on Vaihingen, Germany \cite{cramer2010dgpf}. This additional test aims to demonstrate its effectiveness on complex environments, and to compare it with other existing methods.
	
	We use the true orthophoto mosaic given as a RGB image of 9-cm resolution generated from aerial images that are taken between July 24 and August 06, 2008; whereas the airborne LiDAR dataset of an average point density of 4 points/m$ ^2 $ is acquired on  August 21, 2008. Since the misalignment between the  orthophoto and the airborne LiDAR data was already small, a registration between the datasets was not carried out.
	
	This benchmark data set provides three test areas consisting of buildings with diversified characteristics, as well as their ground truth boundaries. Area 1 is situated in the center of the city and characterized by dense construction consisting of historic buildings with complex shapes and some trees. Area 2 is composed of high-rise residential buildings surrounded by trees. Finally, Area 3 is purely residential with detached houses and many surrounding trees. 
	
	\subsection{Evaluation of snake models }
	
	Fig. \ref{fig:compare_snakes2} presents the results of different snake models applied to extract buildings from the Quebec City data set: basic snake without GVF and $ E_\mathrm{img} $, GVF snake, snake model of \cite{guo2002snake}, snake model of \cite{kabolizade2010improved}, and our proposed method. 
	In this comparison, all snake models are initialized by the LiDAR  building boundaries projected on the optical image (i.e. $ \mathbf{T}B_i $). As we focus on an unsupervised approach for snake models without a priori information of building gray-level, method of  \cite{ahmadi2010automatic} is not considered. 
	Visual comparison shows that our proposed snake model always converges better toward the building true boundaries than other snake models on all presented buildings. 
	
	
	Table \ref{tab:compare_snakes} summarizes the IoU metrics computed based on snake models with respect to ground truth building boundaries. 
	All the snake models except the basic snake model show a high average IoU value, i.e. more than 88 \%, which demonstrate the gain of using the projected LiDAR building boundaries as initial points for snake models. 
	
	However, we notice unstable IoU metric values from GVF snake and snake model of \cite{kabolizade2010improved}, underlined by the IoU minimum values. Furthermore, IoU value of snake model of \cite{kabolizade2010improved} worsens if the building has similar color to its background (cf. Fig. \ref{sfig:c} and \ref{sfig:e}). On the other hand, snake model of \cite{guo2002snake} yields good IoU but it is usually unable to converge toward building corners and boundary concavities (cf. Fig. \ref{sfig:c}-\ref{sfig:e}). 
	

	\begin{table}[h]
		\centering
		\scalebox{0.87}{%
			\begin{tabular}{cccc}
				\hline
				\multirow{2}{*}{Snake models} & \multicolumn{3}{c}{IoU}\\
				\cline{2-4}
				& Mean & Min & Max \\\hline
				
				Basic snake & 61.45 \% & 36.42 \% & 72.61 \%\\\hline
				GVF snake &  88.55 \% & 58.10 \% & 97.57 \%\\\hline
				(Guo and Yasuoka, 2002) & 89.04 \% & 79.85 \% & 96.83 \%\\\hline
				(Kabolizade et al., 2010) & 88.32 \% & 57.68 \% & 97.52 \%\\\hline
				(Proposed model) &90.36 \% & 74.23 \% & 97.74 \%\\\hline
		\end{tabular}}
		\caption{Performance of snake models initialized by $  \mathbf{T}B_i  $ on Quebec City data set, based on IoU metric measured between snake models and ground truth building boundaries.}
		\label{tab:compare_snakes}
	\end{table}
	
	

	\subsection{Overall performance assessment}
	\label{ssec:overall}
	This sub-section is dedicated to the evaluation of the overall performance of our proposed method on both data sets. In addition, using the Quebec City data set, we compare it with the approach using only LiDAR data, in order to demonstrate the benefit of using jointly imagery data with LiDAR data. Then, using the ISPRS Vaihingen data set, we compare the result provided by our method with six existing non-snake building extraction methods.
	
	\subsubsection{Performance results using Quebec City data set}
\begin{figure}[t]
		\centering
		\includegraphics[trim=2cm 1cm 1.75cm 1.25cm,clip,width=0.495\linewidth]{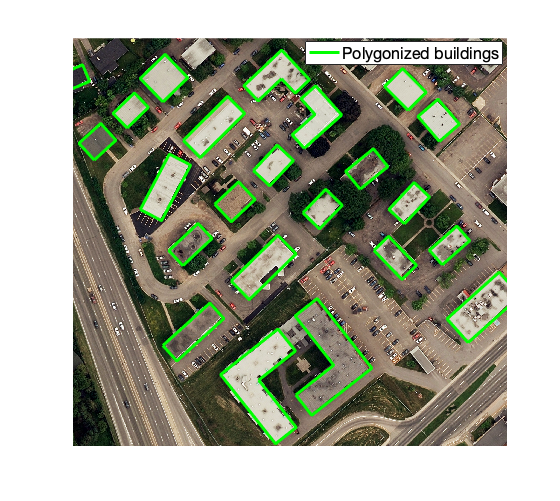}
		\hspace{0.01cm}
		\includegraphics[trim=3cm 1cm 3cm 1.5cm,clip,width=0.485\linewidth]{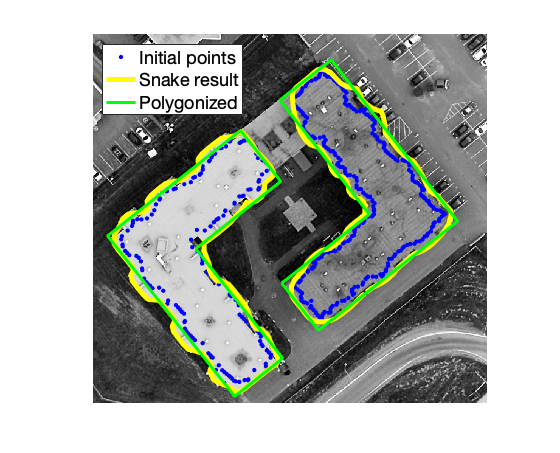}
		
		\includegraphics[trim=2cm 0.5cm 2.5cm 0cm,clip,width=0.49\linewidth]{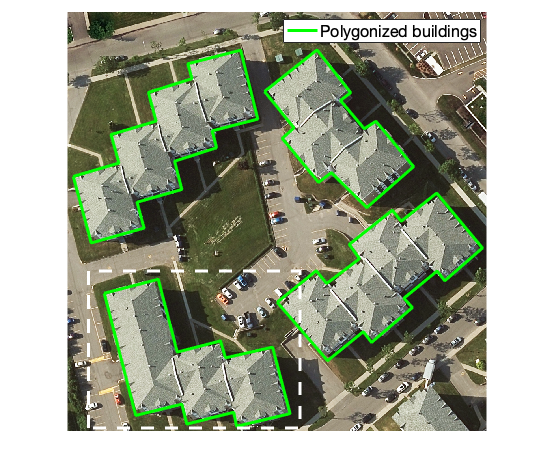}
		\hspace{0.01cm}
		\includegraphics[trim=2cm 0.5cm 2.5cm 0cm,clip,width=0.49\linewidth]{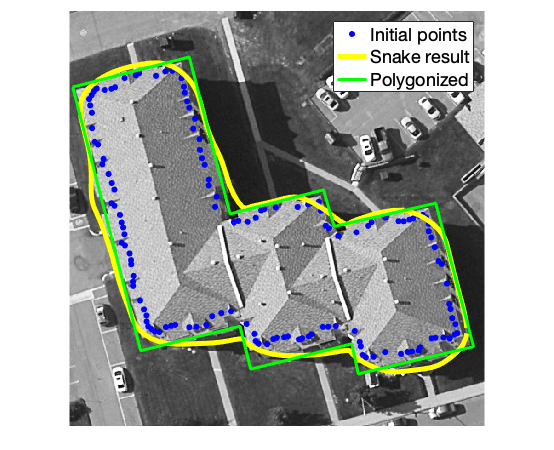}
		\caption{Building extraction result on urban area (Quebec City).}
		\label{fig:BE_by_snake_reg2}
	\end{figure}

\mbox{}\\
	Building extraction result yielded by our proposed method on urban area of Quebec City can be visually assessed through Fig. \ref{fig:BE_by_snake_reg2}. Snake initial points and snake result (before polygonization) are also depicted.
	Many buildings with varying color, or color similar to the background or gable roofs are well delineated. The metric values averaged on the area are summarized by Table \ref{tab:accuracy}. 
	
	%
	
	\begin{table}[h]
		\centering
		\scalebox{0.87}{%
			\begin{tabular}{ccc}
				\hline
				Metric & LiDAR only & Proposed method \\\hline
				
				{IoU} & 85.51 \%& 91.12\% \\\hline
				$ \mathrm{C}_p $ & 87.83 \% & 97.07 \% \\\hline
				$ \mathrm{C}_r $ & 97.05 \% &  92.88 \% \\\hline
				$ \mathrm{EDC} $ & 1.48 m & 0.89 m\\\hline
				{DARE} & 0.62$^\circ $ & 0.81$^\circ $\\\hline
		\end{tabular}}
		\caption{Average pixel-based building extraction accuracy metrics on Quebec City data set.}
		\label{tab:accuracy}
	\end{table}
	
	Compared with metrics of {LiDAR-only method},  
	the proposed method yields more accurate IoU and EDC metrics.
	However, it can be remarked that the value of \textit{Correctness} $ \mathrm{C}_r $ of the  {LiDAR-only} method is higher than the proposed method, i.e. 97.05\% versus 92.88\%. The reason is because in many cases, LiDAR boundary points can be found inside the true building boundaries, e.g. Fig. \ref{fig:initial points}. These cases yield a $ \mathrm{C}_r $ value equal to 100\% (since $ E\cap R=E $ when $ E \subseteq  R $), which increase the average value of $ \mathrm{C}_r $. Also, the DARE value of the proposed method is relatively small which confirms its reliability and accuracy in determining building orientation.
	

	
	\subsubsection{Performance results using ISPRS Vaihingen data set}

	\begin{figure}[!t]
		\centering

		\subfloat[Area 1]{\includegraphics[trim=11cm 6cm 9.5cm 5cm,clip,width=0.5\linewidth]{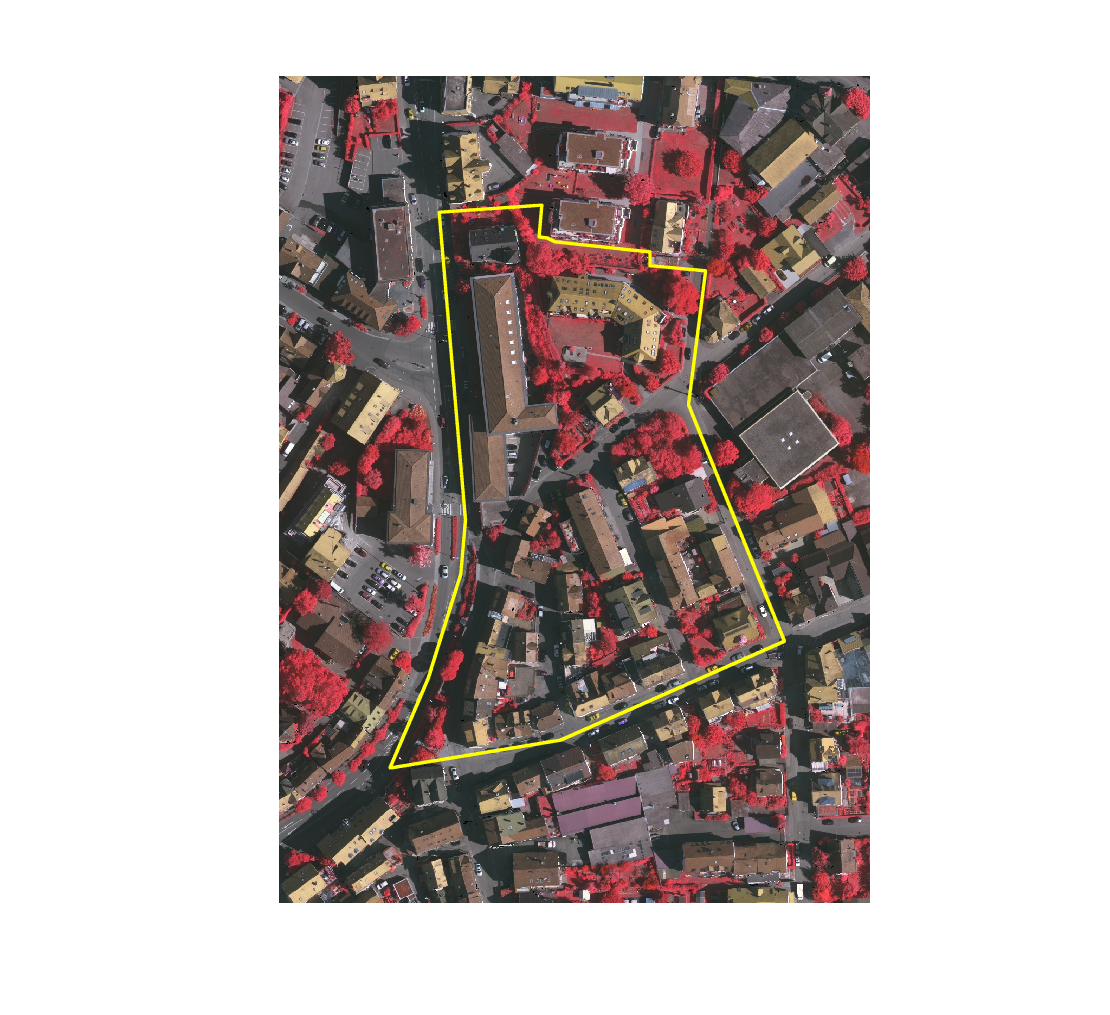}\label{subfig:area1}}
		\subfloat[Result on Area 1]{\includegraphics[trim=5cm 0cm 5.5cm 1cm,clip,width=0.4\linewidth]{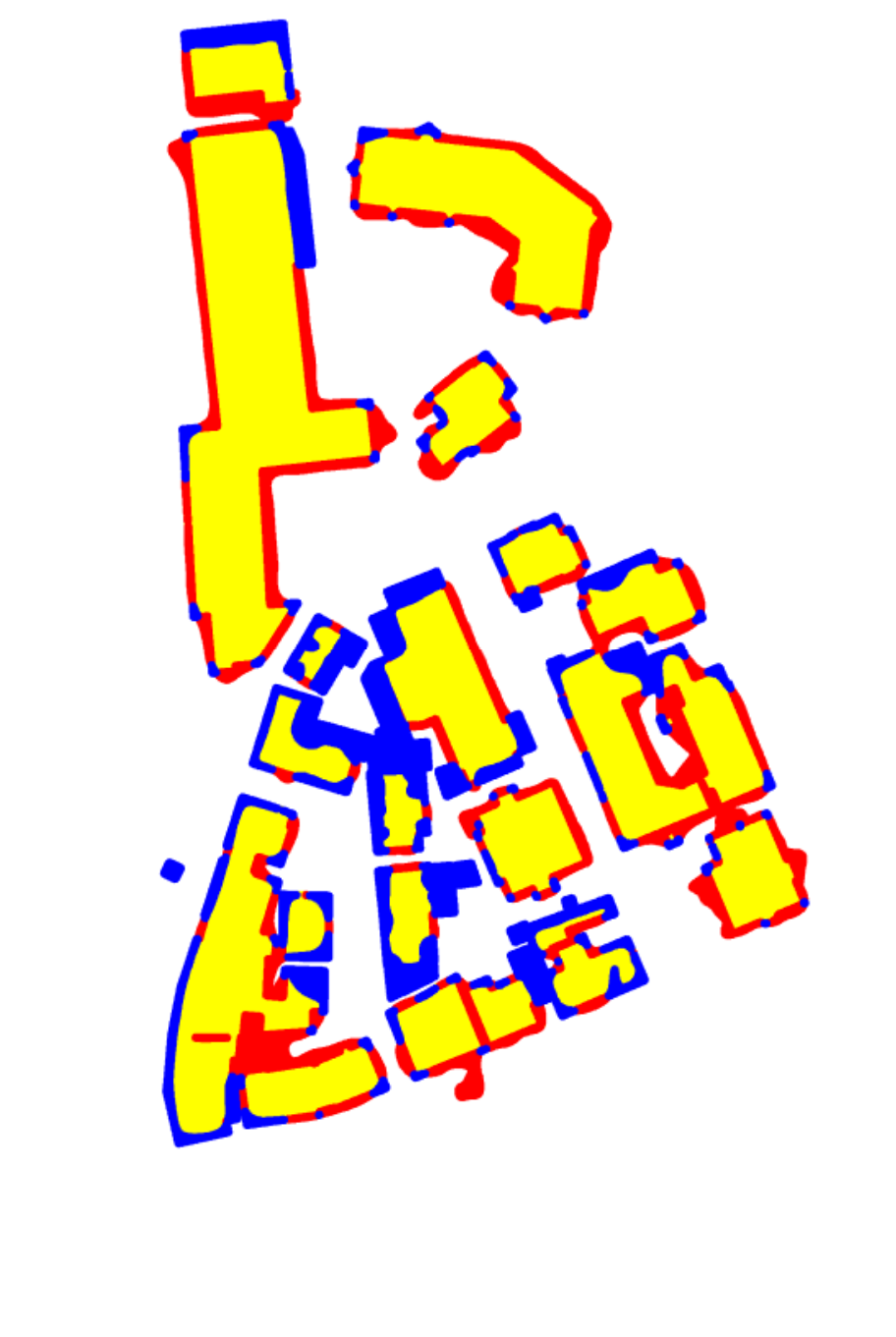}\label{subfig:resarea1}}
		
		\subfloat[Area 2]{\includegraphics[trim=8.5cm 5cm 8.5cm 5cm,clip,width=0.5\linewidth]{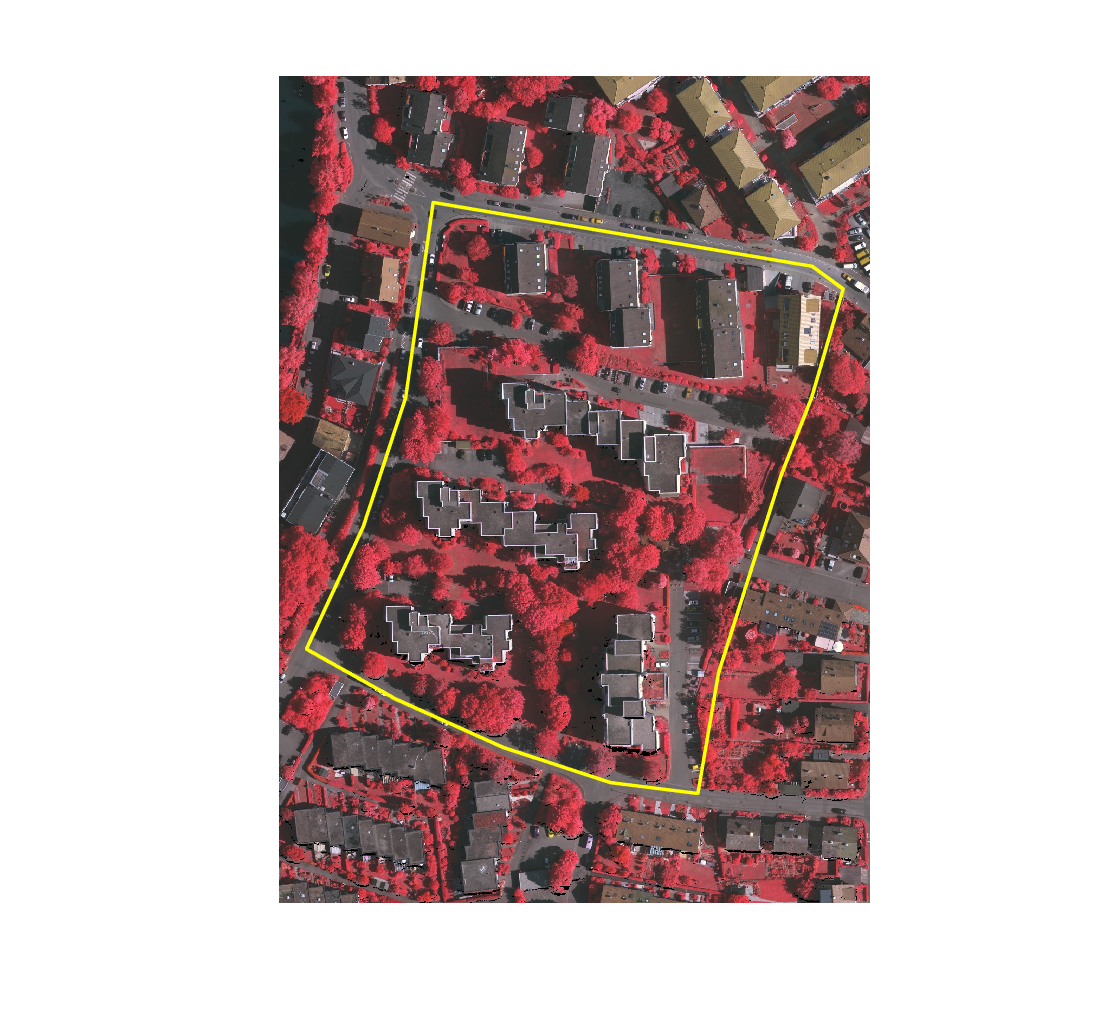}\label{subfig:area2}}
		\subfloat[Result on Area 2]{\includegraphics[trim=4cm 1cm 4cm 0.cm,clip,width=0.4\linewidth]{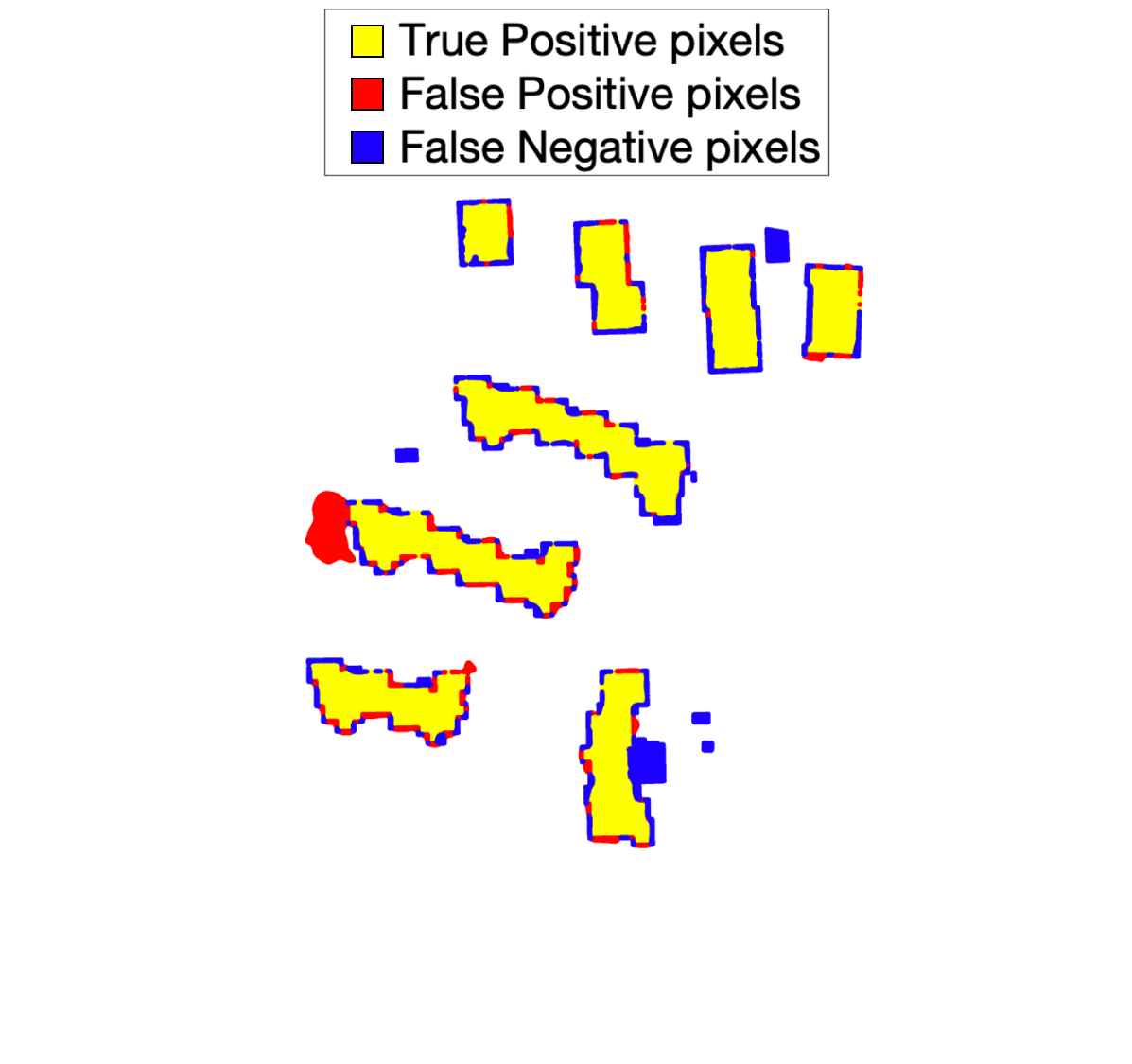}\label{subfig:resarea2}}
		
		\subfloat[Area 3]{\includegraphics[trim=8.5cm 5cm 8.5cm 5cm,clip,width=0.5\linewidth]{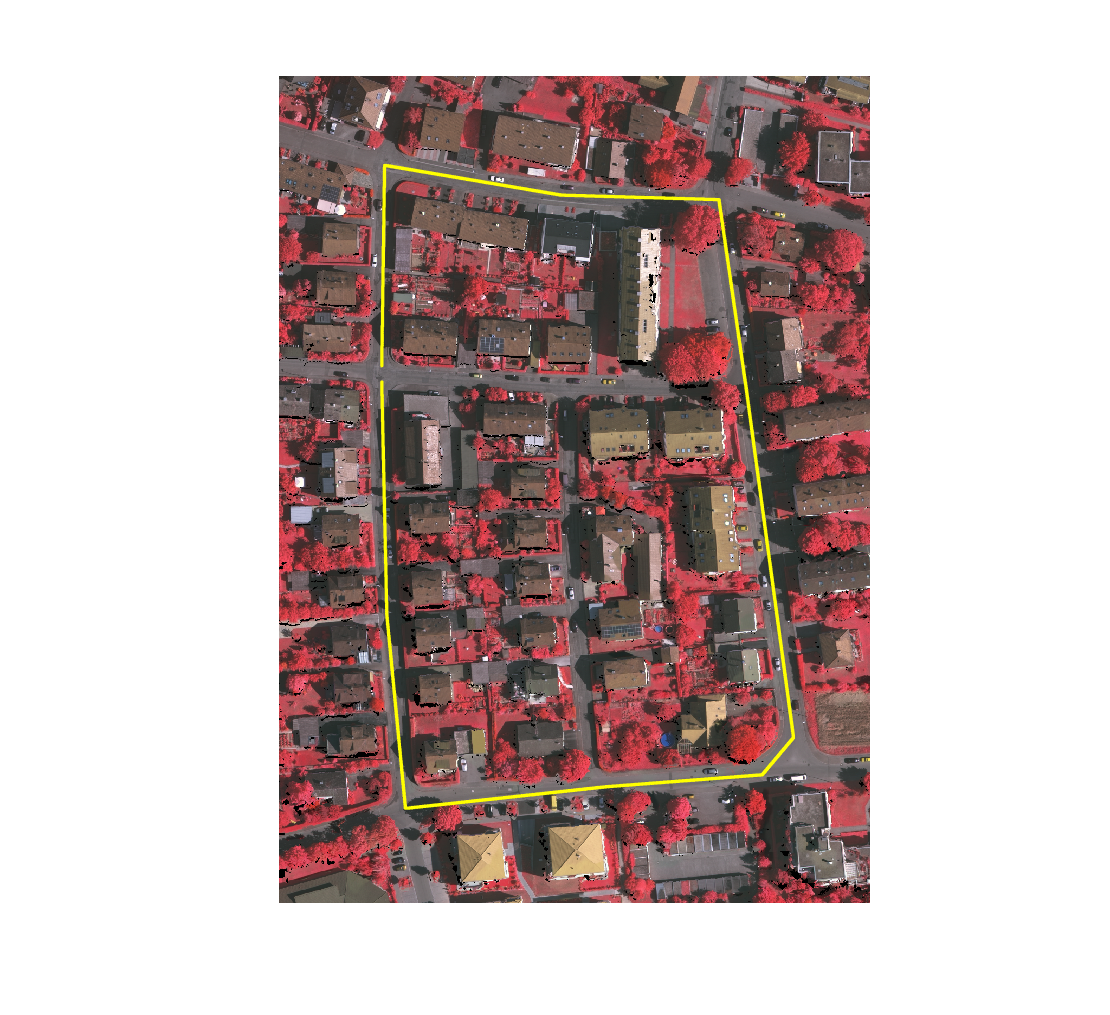}\label{subfig:area3}}
		\subfloat[Result on Area 3]{\includegraphics[trim=5cm 0cm 5cm 1cm,clip,width=0.4\linewidth]{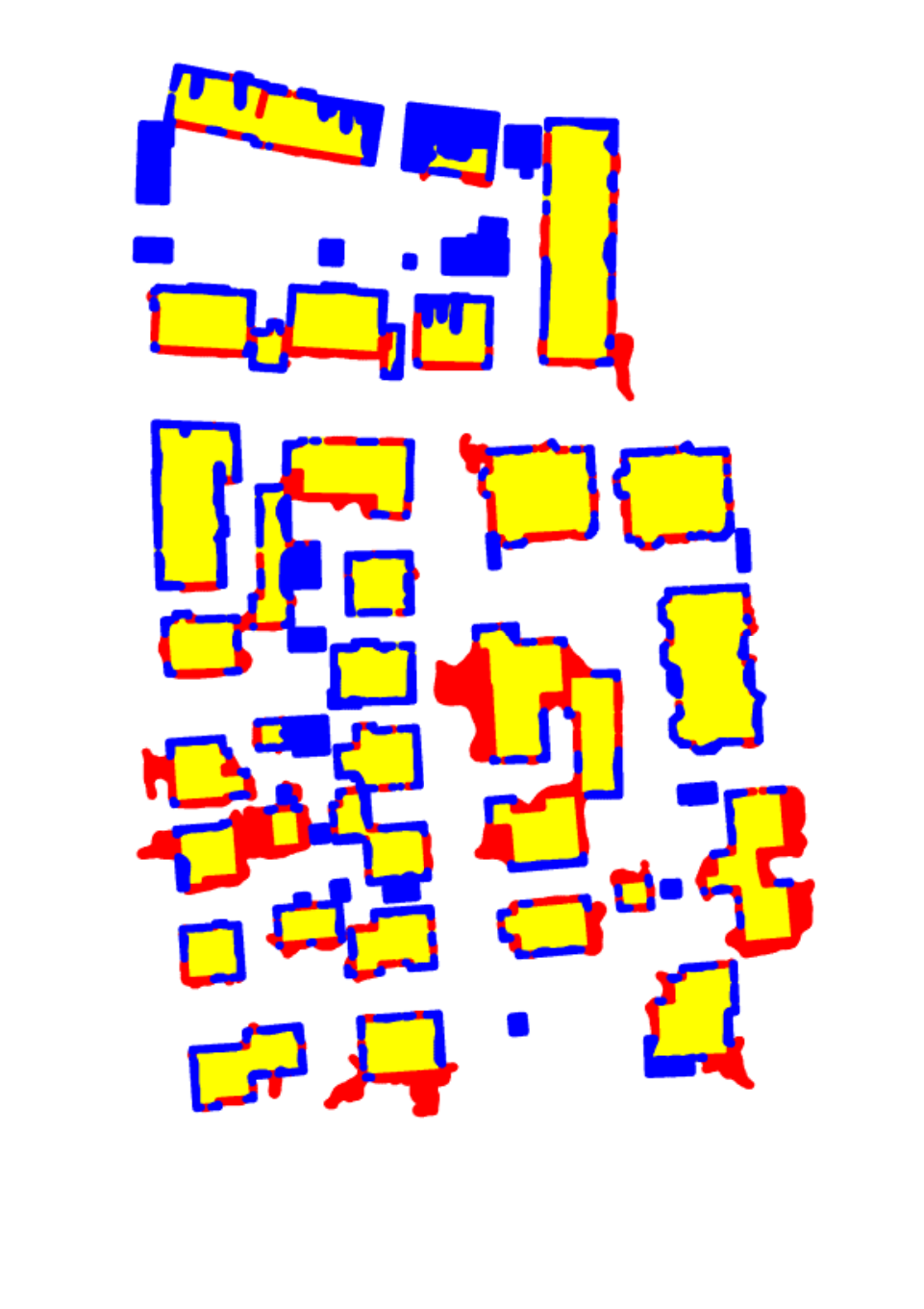}\label{subfig:resarea3}}
		
		\vspace{-0.25cm}
		\caption{Evaluation of building extraction using our proposed method with respect to the ground truth building boundaries on test areas of the ISPRS Vaihingen benchmark data set.}
		\label{fig:isprs_areas}
	\end{figure}
	
	The proposed method is applied  on all three test areas, respectively presented by Fig. \ref{subfig:area1}, \ref{subfig:area2} and \ref{subfig:area3}. Extraction results are then evaluated according to the ground truth building boundaries, provided within the benchmark data set. 
	Fig. \ref{subfig:resarea1}, \ref{subfig:resarea2} and \ref{subfig:resarea3} illustrate this evaluation denoting true positive, false positive and false negative pixels. Their significations were mentioned in sub-section \ref{ssec:metric}.

	First, the results are visually assessed. Although a high percentage of building pixels are well extracted (represented by yellow pixels), several issues can be identified in all three test areas. Firstly, several buildings are not entirely extracted but only partially, causing false negative pixels. This issue is due to  complex building roofs and connected buildings. Also, several small buildings are also not extracted. Secondly, several vegetation regions that are close to buildings are also extracted as buildings, causing false positive pixels.
	
	
	\begin{table}[h]
		\centering
		\scalebox{0.9}{%
			\begin{tabular}{cccc}
				\hline
				Area & IoU & $ \mathrm{C}_p $ & $ \mathrm{C}_r $ \\\hline
				1 & 82.27 \% & 89.43 \% & 91.13 \% \\\hline
				2 & 87.35 \% & 92.13 \% & 94.39 \% \\\hline
				3 & 80.07 \% & 87.21 \% & 90.72 \% \\\hline
				Avg & 83.23 \% & 89.59 \% & 92.08 \% \\\hline
		\end{tabular}}
		\caption{Pixel-based building extraction accuracy metrics on ISPRS Vaihingen benchmark data set.}
		\label{tab:ISPRS_accuracy}
	\end{table}
	
	Pixel-based assessment metrics on each area are provided by Table \ref{tab:ISPRS_accuracy}.  These metrics are then compared with other methods that are selectively proposed by \cite{gilani2016automatic}, and synthesized in Table \ref{tab:methods}. The first two methods are categorized \textit{supervised}, as they involve a supervised classification methodology based on training data, according to the taxonomy by \cite{rottensteiner2014results}. On the other hand, the others proposed \textit{unsupervised} and \textit{data-driven} approaches which consist in extracting buildings without predefined constraint on the building appearance \cite{gilani2016automatic}.
	
\begin{figure}[t]
		\centering
		\includegraphics[width=\linewidth]{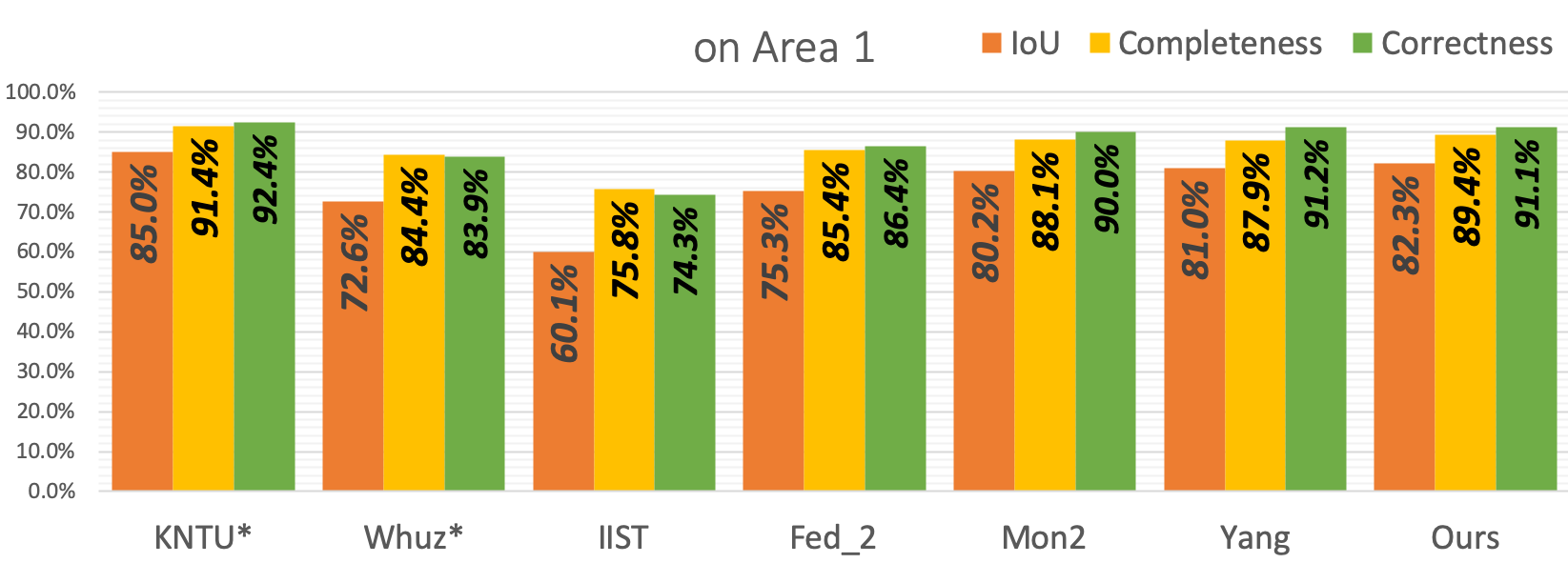}
		
		\includegraphics[width=\linewidth]{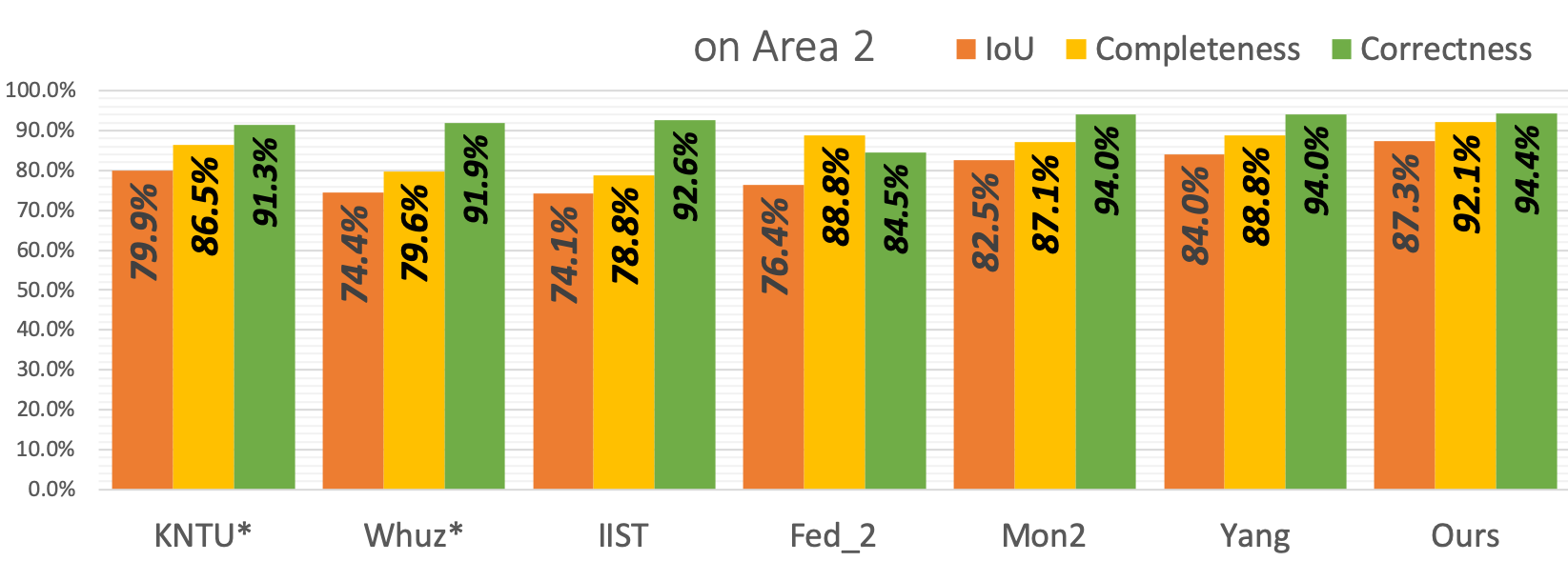}
		
		\includegraphics[width=\linewidth]{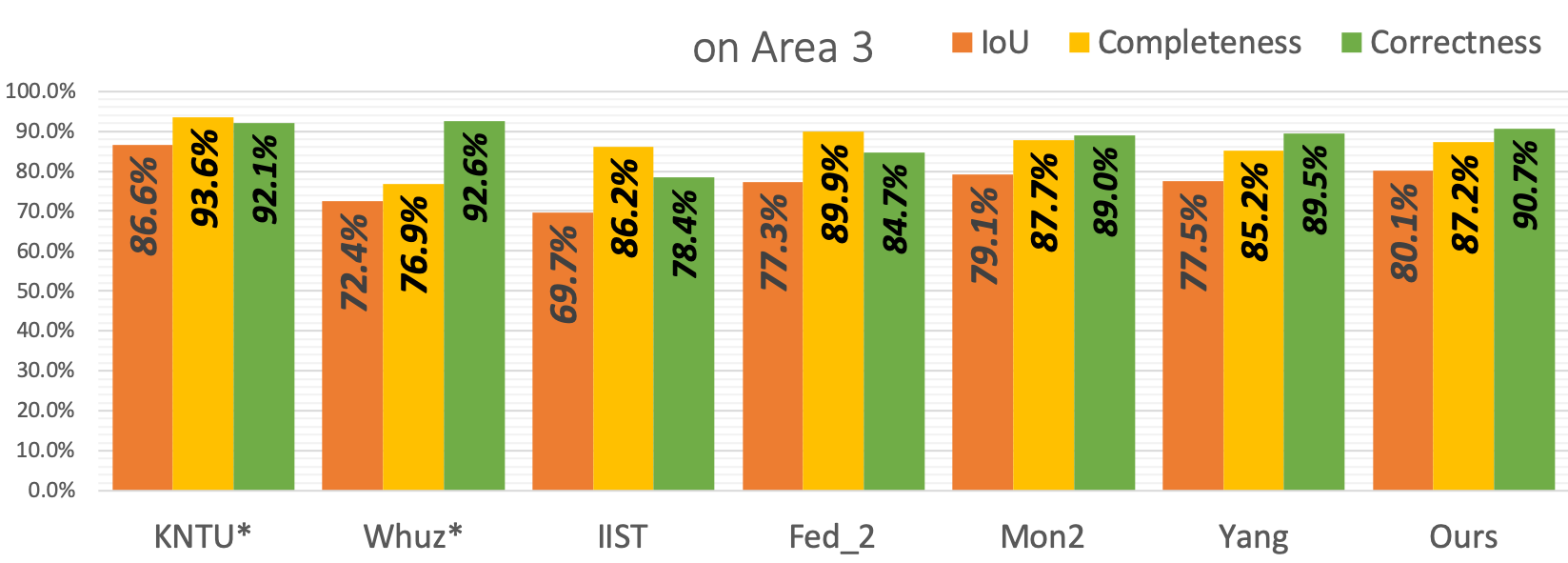}
		\caption{Comparison of building extraction accuracy metrics on ISPRS Vaihingen benchmark data set among various methods. The \textquoteleft $\ast$\textquoteright~sign indicates that a method is supervised.}
		\label{fig:accuracy_methods}
	\end{figure}

	Fig. \ref{fig:accuracy_methods} depicts graphically the performance of each method with relation to IoU, Correctness and Completeness metrics. 
	Despite the discussed issues on this data set, our proposed method, being automatic and unsupervised, shows very competitive quantitative results. Indeed, it achieves relevant accuracy results on all three test areas. For instance, on Area 1 and 3, our proposed method yields better IoU than most of the other methods, except the supervised method KNTU, or on Area 2 where ours exceeds all the others.
	
	\begin{table}[h]
		\centering
		\scalebox{0.9}{%
			\begin{tabular}{p{3.5cm}|c|c}
				\hline
				Method & Data types & Processing strategy \\\hline
				KNTU \cite{zarea2015novel} & \multirow{5}{1.8cm}{\centering LiDAR + image} & \multirow{3}{*}{supervised} \\\cline{1-1}
				Whuz \cite{zhan2012ground} &  &  \\\cline{1-1}\cline{3-3}
				IIST \cite{gilani2016automatic} &  & \multirow{5}{2.5cm}{\centering unsupervised and data-driven}  \\\cline{1-1}
				Fed\_2 \cite{gilani2016automatic}  &   & \\\cline{1-2}
				Mon2 \cite{awrangjeb2014automatic} & \multirow{3}{*}{LiDAR} & \\\cline{1-1}
				Yang \cite{yang2013automated}  &  & \\\hline
		\end{tabular}}
		\caption{Methods that have been compared with the proposed method, using ISPRS Vaihingen data set.}
		\label{tab:methods}
	\end{table}
	
\begin{figure}[t]
		\centering
		\subfloat[Building ground truth (cyan) and extracted boundary (green).]{\includegraphics[trim=11cm 3cm 11cm 3cm,clip,width=0.48\linewidth]{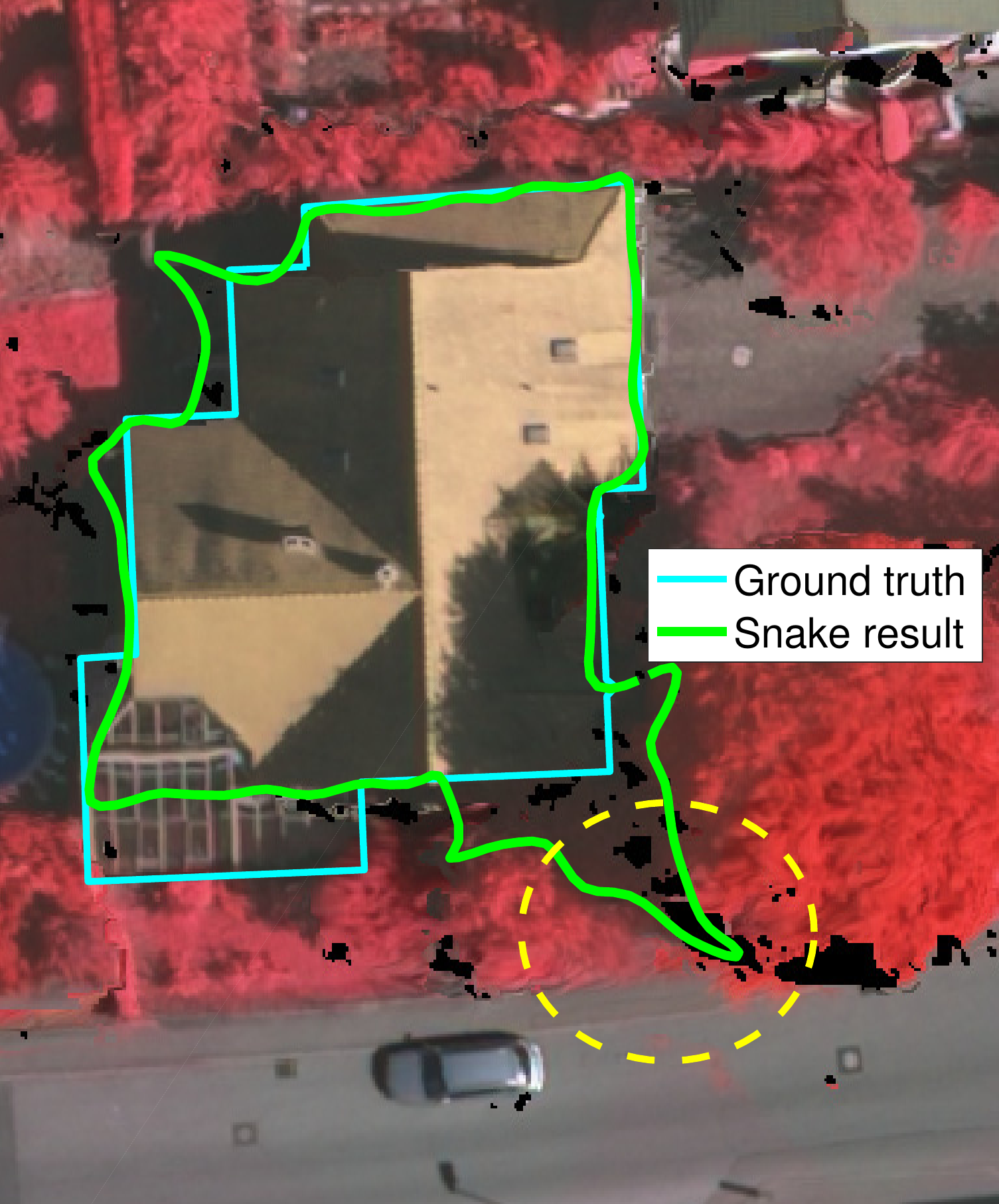}}\hfill
		\subfloat[Binary mask denoting  null-valued pixels (white: null, black: non-null).]{\includegraphics[trim=11cm 3cm 11cm 3cm,clip,width=0.48\linewidth]{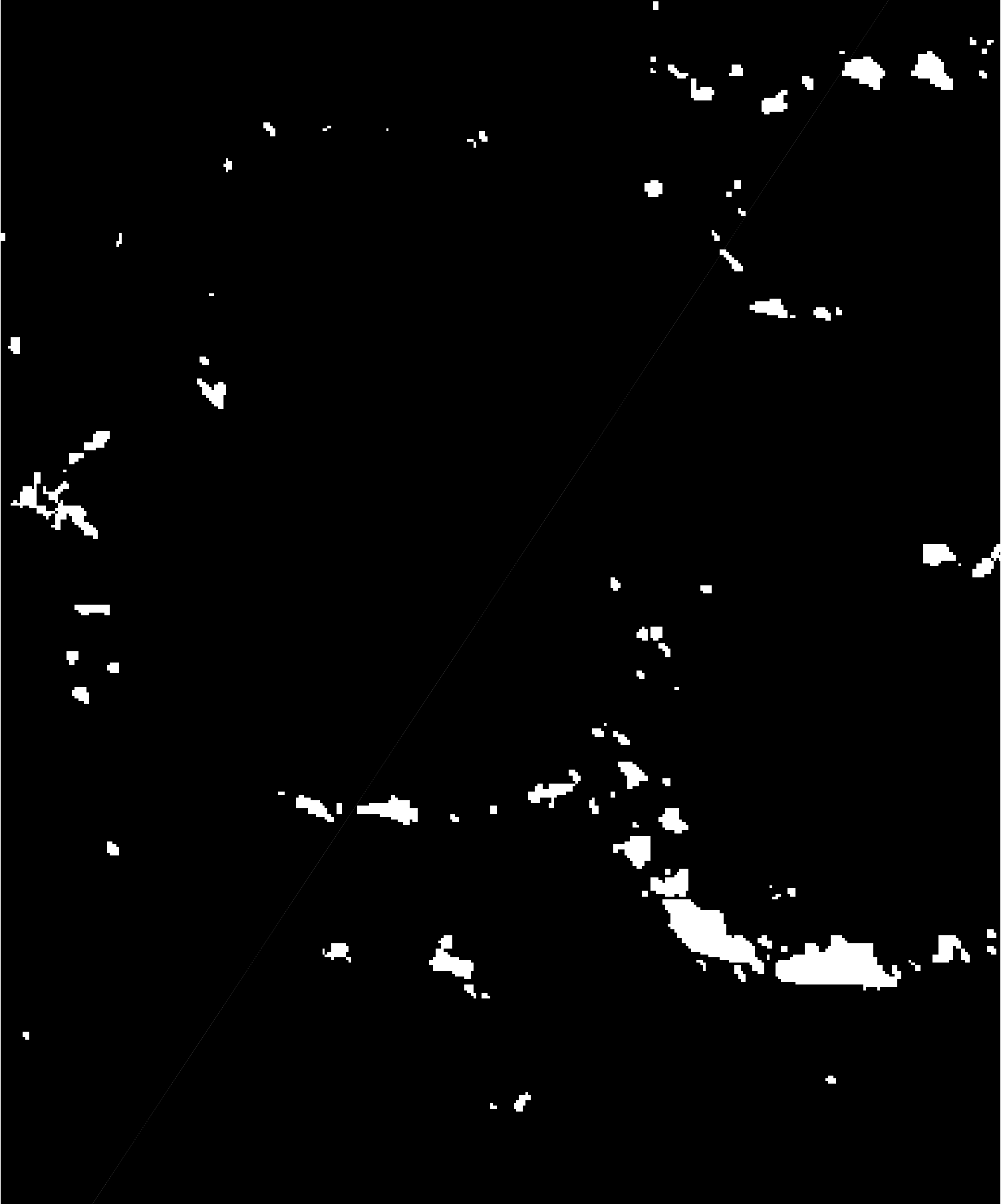}}
		\caption{Illustration of a wrongly extracted building due to null-valued image pixels.}
		\label{fig:hole_img}
	\end{figure}

	However the tests conducted on the ISPRS data set underlined a limitation in our approach. Indeed, there are several null-valued regions in the orthophoto as shown in Fig. \ref{fig:hole_img}. These regions perturb the snake model convergence as they involve high gradient values in the image-based external energy term that attract the snake model.

	\section{CONCLUSIONS AND PERSPECTIVES}\label{sec:conclusion}
	In this paper, we proposed and evaluated an unsupervised method to extract buildings from urban scenes, using snake model on aerial optical imagery with  airborne LiDAR data. 
	Without needing manual initial points nor training data, the proposed method is highly automatic and achieves a high accuracy of building extraction, indicated by different metrics on many test areas (cf. Tables \ref{tab:accuracy} and \ref{tab:ISPRS_accuracy}). Indeed, compared to other methods dedicated to building extraction, our work provide an alternative solution that is unsupervised and yields a better accuracy than most of them, except the supervised method proposed by \cite{zarea2015novel}. 

	These results also demonstrate significantly the advantage of the conjoint use of optical imagery with LiDAR data, even with a LiDAR dataset collected  a long time before, e.g. five years. 
	Indeed, by the virtue of LiDAR-based initialization and guidance, our proposed method succeeds to extract buildings with gable roofs and/or having varying color, or even when the roof has very similar color to its background.	It also overcomes the problem of disturbed shadow regions near buildings which is usually problematic to a building extraction method, e.g. \cite{fazan2013rectilinear}. Compared with methods that use only LiDAR data, the proposed approach provides more accurate building extraction results. Moreover, the fact that this approach uses an optical image with a LiDAR dataset that is acquired several years in advance also means a cheaper and timelier solution, instead of requiring updated LiDAR data.  
	
	However, like other snake model-based building extraction methods, our proposed method is also still dependent to snake parametrization 
	in order to work well on complex areas. Future work will investigate on an automated contextualization of scene (either residential, industrial, or mixed) allowing to automatically parametrize the  snake model. Nevertheless, more efforts will also be put on increasing effectiveness of the snake model on complex environments, such as the areas from the ISPRS benchmark data set. As a matter of fact, the results obtained on Vaihingen data set, even though being relevant compared to other  methods, can still be improved in order to reach performances similar to those of Quebec City data set, i.e.  average IoU values respectively  of $ 83.23 \% $ versus $ 91.12 \% $. There are two reasons that explain this difference. The process of building boundary extraction from the LiDAR point cloud of Vaihingen data set is less effective, due to the misclassification problem of vegetation and buildings. LiDAR points from Quebec City data set have a class value which facilitate the discrimination of building regions  from trees. The second reason is the problem of null-valued regions in the orthophoto as mentioned above. Future works will also concentrate on these two issues.

	%

	\section{Acknowledgment}
	This project is funded by the Natural Sciences and Engineering Research Council of Canada and the Brittany region (France). 
	The authors also would  like to thank the Centre G\'{e}oStat (Universit\'{e} Laval, QC, Canada), as well as Quebec City, Communaut\'{e} M\'{e}tropolitaine de Qu\'{e}bec (QC, Canada) for providing the Quebec City data sets used in this work. The Vaihingen data set was provided by the German Society for Photogrammetry, Remote Sensing and Geoinformation (DGPF) \cite{cramer2010dgpf}.

	{
		\begin{spacing}{1.17}
			\normalsize
			\bibliography{reference} 
		\end{spacing}
	}

\end{document}